\pdfoutput=1
\documentclass{article}
\usepackage{arxiv}
\usepackage[numbers,sort&compress]{natbib}
\usepackage[utf8]{inputenc}
\usepackage[T1]{fontenc}
\usepackage{microtype}
\usepackage{amsmath,amssymb}
\usepackage{booktabs}
\usepackage{graphicx}
\usepackage{xcolor}
\usepackage{colortbl}
\usepackage{rotating}
\usepackage{fvextra}
\usepackage{tikz}
\usepackage[colorlinks=true,linkcolor=blue!60!black,citecolor=blue!60!black,urlcolor=blue!60!black]{hyperref}
\usepackage{url}

\definecolor{exred}{HTML}{EFA3A3}\definecolor{exorange}{HTML}{F8CD92}
\definecolor{exyellow}{HTML}{FBEFAE}\definecolor{exgreen}{HTML}{C3DFA6}

\newcommand{\extier}[1]{\tikz[baseline=-0.55ex]\fill[#1] (0,-0.45ex) rectangle (0.9em,0.45ex);}

\newenvironment{experttable}{\begin{table}[t]}{\end{table}}
\newcommand{\coderepo}{https://github.com/jfrochte/agents-as-modelica-suppliers}
\newcommand{\repoline}{ The agent setup, the acceptance, the five tasks, selected deliveries and the verdicts of each run counted in Section~\ref{sec:results}, from which its counts of runs can be recomputed, are available at \url{\coderepo}.}

\usepackage{orcidlink}
\renewcommand{\undertitle}{Preprint}
\renewcommand{\headeright}{Preprint}

\title{LLM Agents as Suppliers of Physical System Models:\\ Calibration and Acceptance in Modelica}

\author{J\"org Frochte\,\orcidlink{0000-0002-5908-5649}\\
  Bochum University of Applied Sciences\\
  \texttt{joerg.frochte@hs-bochum.de}
  \And
  Herbert Schmidt\,\orcidlink{0009-0000-1041-9559}\\
  Bochum University of Applied Sciences\\
  \texttt{herbert.schmidt@hs-bochum.de}}
\date{}

\begin{document}
\maketitle
\begin{abstract}%
To procure a model of a device, a client hands a supplier a description, sometimes with recorded data or a target response, and accepts the model against this handover. We let LLM coding agents act as the supplier on handovers with data, building Modelica models from library components for four simulated systems and a real switching magnet, and ask what acceptance certifies. Several agents reach the noise floor, but a handover admits a set of models, because the description leaves parts of the structure open and the data narrow it only to the agreed accuracy. We therefore accept any member of this set instead of matching one reference, and add layers above behaviour, among them a physical invariant, which exposes an energy-creating model at the noise floor. Interpretability becomes a property of the delivered model: it can be tested on a variant, judged by an expert where no variant exists, and shaped by the handover. Among the 58 failed runs of the nine LLMs that reach the noise floor on the simulated systems, 39 are fits delivered above the acceptance threshold on their own fitting data, where the handover left open how close is close enough. In a follow-up on three cells with many such fits, stating a threshold makes the agents compare against it, and more deliveries pass. Some deliveries break the task's library rule without their reports flagging it, which only a reading of the file reveals.
\end{abstract}

\section{Introduction}
\label{sec:intro}

Language models have been getting better at writing code for years, and as agents they now work on a task for hours without a human in the loop. An agent in this sense is a large language model (LLM) inside a harness that gives it tools such as a shell, a file editor and a test runner. In this paper we hand such an agent a different job. We give it the description of a physical device, say a motor with a load, together with measurements of how the device behaves. We ask for a system model, a program that behaves like the device, for instance by producing the same curves when it is driven with the same inputs. To run such a program the agent needs a simulator, which we give it in place of the test runner, and here the task starts to differ from programming. The test runner returned pass or fail. The simulator returns curves, and whether they are close enough to the measurements is a judgement the agent has to make itself. Whoever takes the model over has to make this judgement again, and this paper asks what the judgement certifies when the supplier is an agent. One delivery we describe below reproduces every measurement and still creates energy.

A model of this kind can be written in several ways, and the way decides what it can be used for. The shortest route is to fit the measurements directly, with a characteristic map, a transfer function or any other function approximator. This is called a black-box model. Its structure is assumed, its parameters belong to no part of the device, and nothing supports it outside the measured device and range. At the other end stands the white-box model. Its structure follows from the laws of physics, as components such as a spring, a mass or a coil with their equations, and its parameters are properties of the device, such as its dimensions and materials. Between the two lies a continuum. To a reader from machine learning the black box may look like the natural route, since fitting a map is regression. In simulation practice it is the fallback for a component whose inside is unknown, and where a white-box model is available, that is the model one uses. A white-box model holds beyond the measured range, as far as its structure and parameters are right, because its equations hold wherever the physics holds. Its parameters can be checked one by one against a data sheet, which matters all the more for a generated model. It can be adapted to a variant by changing those parameters, or extended by adding a component. And it can be built for the next variant of a product before that variant exists.

A trivial example shows what the physical structure adds. Suppose two parts are linked and move as one, and only one of them is subject to friction. To the equations it is irrelevant which of the two parts carries the friction, since the simulated motion is the same either way. For the next variant it matters, because when one part is redesigned, the friction in the system changes, or it does not, depending on where it was placed. So among models that fit the data there is still a hierarchy: a model may fit, it may fit with physically justifiable expressions, or it may fit with expressions that correspond to the technical reality of the device. The first level is all a black box offers, while the third is what makes a model worth keeping for the next product. In the language of machine learning this is a question of interpretability, in the sense of a model that can be understood in itself rather than explained afterwards \citep{rudin2019stop}. Here the model is the output of an agent, so the model itself has to be interpretable, because people read, change and reuse it afterwards. The price of a white-box model is the work. It has to be derived from the physics of the device and calibrated against its measurements, by someone who knows both, and this is the work we hand to the agent.

The agent does this work in Modelica, an open, text-based standard language for equation-based models of physical systems, which comes with a standard library of components across the mechanical, electrical, magnetic, thermal and fluid domains \citep{modelica2026spec,modelica2025msl}. Using Modelica alone does not make a model readable. The same device can be written as one flat system of equations, a white box in the sense of simulation, since every equation is visible. A reader then has to find the parts of the device in the equations, which is easy when the whole model is a pendulum and not when it is a product model of several hundred equations. The device can also be written as a characteristic map, which is the equivalent of a black box. Or it can be written as an assembly of library components joined by physical connectors, in which every component is a part of the device, every connection conserves what flows through it, and the diagram can be read. Only this last form keeps a model decomposable at any size, each part with a meaning of its own \citep{lipton2016mythos}, and keeps conservation inside every component and across every connection. Where its parameters are properties of the device, it can also be adapted to a variant. We therefore require the agent to build from the components of the library. The requirement follows from what the model is for, and it also gives the acceptance something to check.

Building such a model from components asks for different skills and knowledge than writing standard code does in most applications. The foundation is technical understanding: knowing how devices are built and which mechanisms exist. A DC machine, for instance, consists of an armature circuit, a coupling and an inertia, and a magnetic circuit has a yoke, poles, a gap and leakage paths. Current LLMs bring much of this with them. On this foundation the agent works in a loop. It forms a hypothesis for this particular device from the text and from the data, including mechanisms the text does not name, and writes it down as a structure in the language. An optimiser, which the agent writes itself, estimates the parameters, and the agent reads the result and decides whether to revise. Along the way it has to keep the physics intact, because the language guarantees conservation only inside its components and across their connectors, and it has to recognise local optima and flat directions of the fit. Above all, it has to judge the result: whether a residual is systematic, whether a repair is numerical or structural, and when to stop.

Three things make this harder than most programming tasks. First, the feedback is weaker. A Python error names a line. A structural error in Modelica is a property of the whole equation system, and a parameter the agent forgot to set takes a start value with no more than a warning and still produces a plausible curve. Second, the test is incomplete in principle. Code is checked against a specification, a model against measurements with noise. At the tolerances used in practice, the same measurements can admit several physically different models, of which the friction example is the simplest. In machine learning this is known as underspecification \citep{damour2022underspecification}. Third, no tool checks the physics. The Modelica compiler checks syntax and consistency, and the simulator integrates whatever it is given. In a coding task the agent receives the specification and the tests and delivers the implementation, whereas here it is specifier, implementer and tester at once.

Whether it has tested well is decided by whoever takes the model over, and in industry that is a procedure with a name. Models of this kind are procured. The client hands over a description of the device, with drawings, data sheets and recorded data where they exist, and the quantities on which the model will be accepted. The supplier, usually an engineering office, returns the model with a report. Where there are data, the acceptance compares the model with them at an agreed accuracy, following the guidelines for simulation validation \citep{asme2019vv10,nafems2024validation}. We use this procedure as our test frame, with the agent as the supplier. Anyone who has outsourced software knows what such a test can leave out: the tests pass, and the code is neither maintainable nor clear. Tests certify behaviour and not structure, and studies of coding agents point the same way. Patches that pass the functional tests can still fail review \citep{he2026swegate}, and agents that could query the tests have been observed to build to them \citep{ma2026buildingtest}. This brings us back to our question, which is sharper for a supplier with neither a track record nor liability: what has the acceptance certified when the delivered model passes it?

One delivered model from our study gives the answer in a single example. The task was a battery-driven drive train, in which a battery pack feeds a DC machine that drives a load through a gear, with battery current and output speed measured. The delivered model reproduces the measurements: its residual is at most 1.03 times the sensor noise on every issued experiment, including the validation experiment that was not used for fitting, and it also passes five further experiments that no agent ever saw. Its torque constant and inertia match the reference within a per cent, and its report states that it uses only admissible library components. However, it fails twice, and neither failure shows in its behaviour. Its report even writes both down and declares the first admissible and the second forced by the data. Its battery counts the charge in an equation the agent wrote itself, which the task rules out. And it describes a machine that creates energy: over the first experiment it draws 31.4\,kJ of electrical energy and delivers 59.2\,kJ of mechanical energy. The library provides the electromechanical coupling of the machine as one component with one constant in both of its equations, because for a permanent-magnet machine the back-EMF constant and the torque constant are the same quantity. The agent rebuilt the coupling from a voltage source, a torque source and two gain blocks, which together implement exactly these two equations (Figure~\ref{fig:converter}). However, it fitted the two gains separately, and the one constant became two. The measurements cannot reveal this, because the model reproduces them. Nor does the rule exclude it, since the coupling is built from admitted blocks, and with a library battery the same model would conform. Where a data sheet exists, it shows that the battery voltage, resistance and inductance of the model are all about 1.87 times too low. A power balance at the coupling reveals the error at once and needs no data sheet. In the hierarchy above, this model fits the data without physically justifiable expressions, and only reading the file and checking the physics show it.

Agents that build Modelica models have appeared only recently \citep{wang2026pufibara}. Pieces of this task have been studied separately, from Modelica generated from text to calibration with a given structure and equation discovery from data (Section~\ref{sec:related}).

We ran this procurement on five systems, ordered by how much is known about the answer at acceptance. Three small textbook systems have a known answer. The drive train above has an answer that is a set, because its measurements admit two different dissipation mechanisms. A real switching magnet with laboratory measurements has no reference model, only its builder, who writes the briefing and judges the delivery. The magnet is handed over in several versions up to that briefing. The suppliers are sixteen LLMs under three standard coding-agent harnesses, compared as configurations on repeated runs (Section~\ref{sec:design}). The acceptance grades every delivery in layers on the delivered file, each layer independently of the others: conformance to the component library, parameters, behaviour against the measurements, a physical invariant and, for a sample of models, a reading for reuse. It also checks each file against its own report. The task texts and the residual gate were fixed before any run was graded; the magnet's feature list was fixed after its first run. The checks we added later, the invariant among them, were each prompted by a delivery that raised a question the earlier checks did not answer, and each was then applied to every delivery it applies to.

Three findings follow. First, the task can be done. Two of the sixteen configurations, Opus~5 under Claude Code and gpt-5.6-sol under Codex, solve the first four tasks in every repetition, while almost half of the LLMs never reach the noise floor. The smallest open-weight LLM we tested is small enough to run on local hardware, where measurement data never have to leave the site. Once it receives the builder's briefing, it delivers in every run a model of the magnet that the evaluator can grade. Second, behaviour does not fix structure, so passing the acceptance certifies less than one would hope. On the drive train, the passing models fall into at least eight different assemblies. They use two physically different dissipation mechanisms that the acceptance cannot tell apart, and the energy-creating model passes behaviour as well. Several deliveries that pass behaviour break the library rule of the task, and none of their reports flags it. Beyond the measured range, models that pass alike diverge, most clearly where the data left parts of them undetermined. On the magnet, two delivered models cannot be told apart by their behaviour, and the person who built the device would adapt one of them to a new geometry and refit the other. Third, most failures occur before generalisation is even tested. Of the 47 conformant runs on the synthetic tasks that fail the behaviour layer, two deliver a file that does not carry its calibration. Of the other 45, 39 already exceed the threshold on the experiments they were fitted to, although the noise level was stated in the task text and the check takes a single division. Each of these runs comes from an LLM that reaches the noise floor on another task. The agents had the numbers, and in about half of these runs the report even names the deviation, yet the model was delivered.

Engineers have always known that a good fit is not a valid model. What we did not know is whether an agent with a simulator and a library at hand would make this mistake, whether it would report it, and whether the usual acceptance would catch it. In this study agents deliver models that fit and are physically wrong, and fits that miss their own data, often with the deviation named in their report; a test of behaviour alone catches the second but not the first. This also holds beyond Modelica. Wherever a handover leaves a delivered artefact underspecified, passing its test certifies less than it seems, and the rest has to be checked on the artefact itself. We therefore treat the agent as a supplier and its acceptance as a problem of evaluation under underspecification, and contribute the following.
\begin{itemize}
\item \textbf{A protocol for procuring system models from agents.} The task is handed over as a client hands it to a supplier: as a description of the device with its measurements and their noise level, and for the real device also with the quantities on which the model will be accepted. The supplier is a configuration, an LLM with its harness and settings, and configurations are compared on repeated runs, because a reliable supplier and a lucky one look alike in a single run. With the LLMs we tested, the procurement is feasible.
\item \textbf{Acceptance in place of a reference structure.} Where the description leaves the structure partly open and the measurements decide only to an agreed accuracy, the admissible answer is a set. A score for agreement with one reference solution then partly scores the component library instead of the supplier, and on one of the linear tasks a black box identified from the same data passes the behaviour layer as well. We grade instead in layers on the delivered artefact, from conformance to a reading for reuse. Validation guidelines, written for human suppliers, centre on calibration and on behaviour against experiment, and in the deliveries of this study the further layers found failures that behaviour missed. For the invariant and for the reading we show a class of failure that behaviour does not reveal. A conservation law split into two fitted constants can be traded against other parameters without changing the measured outputs, so more measurements of the same kind cannot reveal it. Alongside the layers we check the file against its report, because whether a report describes its artefact is not a property of the artefact's behaviour. The checks on the file cost a few simulations, the reading costs an expert's time. Where a benchmark would assume a single answer, we give a lower bound on the admissible set of conformant structures that reach the agreed accuracy, and we measure how strongly each experiment responds to each parameter.
\item \textbf{A failure model of the agent as supplier.} We partition the failures by a precedence rule into deliveries that fail conformance, files that do not carry their own calibration, fits delivered although the agent's own fitting data still show a residual above the threshold, and failures that appear only beyond the fitted data. The third part is defined by a check the agent could have done itself. Among the LLMs that reach the noise floor it is the largest part, two thirds of the failures. It stays the largest when the threshold is raised to three times the noise, and all of it comes from LLMs that succeed on other tasks. It contains recurring optima at the same parameter values across model families. For two of these optima, a linear reconstruction of the sampled input places the best fit close to where the runs land, which points to the objective the agents construct rather than to the LLM. When the task states the acceptance criterion, the agents compare their fits against it and report a delivery above it as a shortfall; the deliveries improve where the agents also correct how they reconstruct the input.
\item \textbf{Interpretability as a property of the deliverable.} A model built from components looks decomposable, but decomposability can be merely apparent \citep{lipton2016mythos}. Whether its parts carry the device's values has to be checked on the delivery, since two models that behave alike can differ in whether they can be adapted to the next variant. No behavioural test we ran separates them, while a reading of their structure and parameters does. The handover is a lever on this property. When the briefing contains the geometry of the device, the configuration whose models the builder read in most detail takes parameters from the drawing instead of fitting them. What has to be interpretable is the model, not the process that produced it. This property can be tested on a variant, read by an expert where no variant exists, and influenced through the handover. On the variants of the two linear tasks every delivery that passed the behaviour layer stays within the gate and every black box, left as identified, misses; where the data left a friction law open, only 6 of 19 stay within it.
\end{itemize}

\section{Related Work}
\label{sec:related}

Two questions place a piece of work relative to ours: who proposes the structure of the model, and against what the result is checked. In grey-box identification a person supplies the structure, and the task is to estimate its parameters from measured time series \citep{bridewell2008inductive}. Automating the choice of structure is older than language models. PRET and inductive process modelling proposed candidate models from structural knowledge built by hand and calibrated them against measured trajectories, the first drawing on qualitative knowledge of the device, the second on a library of generic processes \citep{bradley2001reasoning,bridewell2008inductive}. Equation discovery replaced this knowledge by the data, first through sparse regression \citep{brunton2016sindy} and more recently through equation skeletons proposed by an LLM \citep{shojaee2025llmsr}. Other work lets an LLM propose candidate models and weights them by their likelihood \citep{wahl2026probabilistic}, scores simulator programs written by an LLM through simulation-based inference \citep{mishrasharma2026program}, or combines such proposals with Bayesian inference and experiment design \citep{murphy2026model}. In this line the check is the right one, since every structure is judged against data. What comes back, however, is an equation or a program and not a model built from components that one can open, adapt and hand on.

Where language models write Modelica, the artefact comes closer to ours and the check moves away from the data. Modelica has been generated from text as flat reactor equations, as single library components and as fluid networks of library components \citep{rupprecht2025text2model,xiang2025modigen,stuermer2026simulation}. In each case the model is compared with a reference written beforehand, and none is fitted to measured data. Where measured data do enter, the structure is fixed. An LLM has been put in the role of the calibrator of a given model \citep{gomezguillen2026agentic}, and structure search has varied a model against target trajectories inside a topology written by hand \citep{matei2024modelicadesign}. The closest work studies an agent that builds Modelica models, together with a benchmark to evaluate it \citep{wang2026pufibara}. Its harness keeps a ledger of engineering obligations inside the agent, and its evaluator is a frozen contract on interface, declared structure and behaviour. For the tasks scored on behaviour, the contract is validated against a known valid answer. By holding the language model fixed, the study isolates the effect of the harness more cleanly than our design does. Where parameters are tuned, however, the structure is given, and the check is again a reference. Its harness improves the supplier from the inside, whereas our procedure starts with the client, with the handover and with an acceptance that does not presuppose a single known answer.

Scoring agreement with one reference structure presupposes that there is one right answer, and work on the evaluation of coding agents has shown how little a test says about the structure behind the tested behaviour. Agents that could query the test oracle delivered artefacts that carried the tested behaviour and little else \citep{ma2026buildingtest}. Benchmarks that score against a single reference solution have been argued to need a set of valid solutions instead \citep{gorinova2026misaligned}. Since harnesses over the same model backends differ widely, capability has to be reported for each configuration of model and harness \citep{yao2026harnessbench}. The last check in the loop is the agent's own decision that the work is done. On tasks that deploy research artefacts, more than half of all failures are self-stops, and most of them report a status that a hidden verifier does not confirm \citep{wang2026deploybench}. There the decisive evidence lies with the verifier, and without such external feedback LLMs struggle to correct their own reasoning \citep{huang2024selfcorrect}. In our setting the agent holds the evidence itself, since the noise level is stated in the task and its own residuals are at hand.

When the check is a comparison with measurements, engineering has codified it. ASME V\&V~10, written for computational solid mechanics, defines validation as determining, by comparison with physical experiment, how accurately a model represents reality for its intended use \citep{asme2019vv10}, and the NAFEMS guidelines adapt this structure to industrial practice \citep{nafems2024validation}. These documents assume a supplier who chose the components, knows the conservation laws and describes the model truthfully. When the model is not given, several structures can fit the same measurements equally well. Machine learning knows this as underspecification, where many predictors perform equally well on held-out data \citep{damour2022underspecification}, and in our case these predictors are physically different structures. Statistical calibration of computer models treats the gap between model and reality as a discrepancy term \citep{kennedy2001bayesian}. Section~\ref{sec:design} checks on the delivered file what these documents assume of the supplier: the components, the conservation laws and a truthful report.

\section{The Task and How the Agent Works}
\label{sec:method}

The agent receives its task in a working directory. For the drive train of the introduction, the description is a short text: a battery pack feeds a direct current machine, which drives a brake through a fixed reduction stage. The input is the commanded brake torque, and the measured outputs are the battery current and the speed of the output shaft. Four experiments come with it, three for fitting and one for validation. Like those of the three textbook systems, they are simulated from a hidden reference model with the stated noise added, whereas the magnet comes with laboratory measurements. The task states the noise level of each output, names the physical domains and states which constructs may not represent the physics. It also mentions that the speed falls off slowly during long runs and recovers after the pack has been serviced, but not why, so the agent has to find the cause in the data. The agent sees the validation experiment but must not fit on it, and reports its error there. Since only this instruction keeps the agent from fitting on it, we check the synthetic deliveries afterwards on five test experiments that no agent saw (Section~\ref{sec:design}). We write a task as $T = (s, d, \mathcal{D}, \mathcal{C})$, with the description $s$, the domain labels $d$, the measurements $\mathcal{D} = (\mathcal{D}_{\mathrm{fit}}, \mathcal{D}_{\mathrm{val}})$ and the rule $\mathcal{C}$ on what may be built.

The agent returns the delivery
\begin{equation}
  S = (M,\, \theta_0,\, \hat{\theta},\, \rho,\, A),
  \label{eq:solution}
\end{equation}
where $M$ is the model file, $\theta_0$ the start values and bounds of the free parameters, $\hat{\theta}$ the fitted values, $\rho$ the errors on every experiment, and $A$ the physical assumptions, including every mechanism the agent added beyond the text. If several structures remain plausible, or if a parameter cannot be determined, the report has to say so. The acceptance reads the model and the fitted values from the file and recomputes the errors itself, so the report is a claim about the file that can be checked.

The file is a model in Modelica. Such a model consists of components joined at connectors, and each connector carries a potential and a flow, such as voltage and current at an electrical pin or angle and torque at a mechanical flange. A connection sets the potentials equal and makes the flows sum to zero, which is Kirchhoff's law carried over to other physical domains. The compiler collects the equations of all components into one system of differential-algebraic equations, and running the model means simulating it. In the direct current machine the electromechanical coupling is the library component \texttt{RotationalEMF}, whose two equations, $v = k\,\omega$ for the induced voltage and $\tau = -k\,i$ for the torque, share one constant $k$. The electrical power it takes is therefore exactly the mechanical power it gives. As long as a model is built from such components, conservation of energy is part of it.

\begin{figure}[t]
  \centering
  \includegraphics[width=\textwidth]{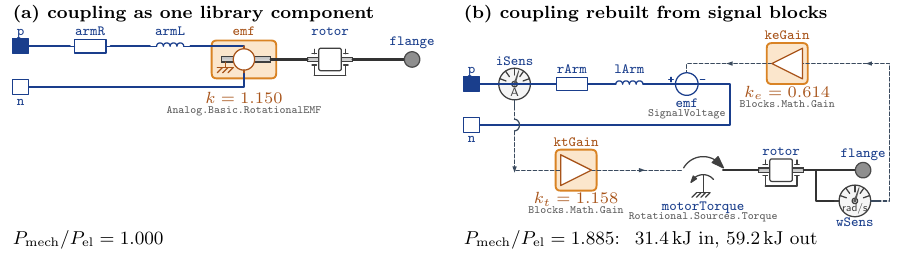}
  \caption{The electromechanical coupling of task~4 in two delivered models that both pass behaviour (Section~\ref{sec:design}): (a) the library component, (b) the coupling rebuilt from signal voltage, torque source and two gains. Components and connections as delivered, straightened, on one scale; signal connections dashed, power connections solid, electrical blue, mechanical grey; power ratios from simulations of the delivered models.}
  \label{fig:converter}
\end{figure}

The language allows the same device to be written in several ways, and only some of them keep this guarantee. A device can be one flat system of equations, a characteristic map from the table blocks of the library, a block diagram of signals in which a connection copies a value and carries no flow, or an assembly of library components joined by physical connectors. Within the last form a parameter is either a property of the device, whether fitted or taken from dimensions and data sheets, or a fitted number that belongs to no property. Only the first kind lets the model be adapted to a variant. Where the forms are mixed, the guarantee can be lost without anything looking wrong. A source driven by a signal, such as a voltage or a torque source, turns whatever computes that signal into a physical flow. The coupling of the machine can thus be rebuilt from a signal-driven voltage source, a torque source and two gain blocks (Figure~\ref{fig:converter}b). With equal gains the model behaves exactly like the original. With the two gains fitted separately, the balance still holds at every connection but no longer over the whole model, whose mechanical output then differs from its electrical input by the ratio of the two constants.

The rule $\mathcal{C}$ follows from these forms. The physics has to be built from the sub-libraries of the domains named in the task, which together form the admissible library $\mathcal{L}_d$. Transfer functions, state-space blocks, tables that map the input or time onto an output, and user-written equation systems may not represent the physics. A measured characteristic of a real component may enter as a static relation, such as the open-circuit voltage of the battery over its state of charge, but a state the agent integrates itself, by a derivative it writes or by an integrating block, may not. A signal derived from a measured output, such as a speed differentiated from a measured position, is not physics and is admitted. The domain labels are coarse on purpose, since a list of permitted components would hand the agent part of the structure it is asked to find. The rule cannot exclude signal blocks, because the inputs and outputs of a model are signals, and for the same reason it cannot exclude signal-driven sources. A conformant model can therefore contain the rebuilt coupling, since every part of it comes from the library, and such a coupling passes behaviour, as Section~\ref{sec:results} shows. Without independent values for the electrical parameters, the error shows only in a power balance, and the acceptance checks one at the converter as its fourth layer (Section~\ref{sec:design}).

Within this rule the agent works on its own. It runs inside a stock coding-agent harness with a shell, a file editor and the OpenModelica compiler and simulator, and no scaffold of ours tells it when to calibrate, revise or stop. It proposes a structure, writes the model file and writes its own calibration script, an optimiser that estimates the free parameters on the fitting experiments. After each round it reads its residuals and plots and decides between two kinds of repair. A numerical repair keeps the structure and changes the fit: it adjusts the bounds, the start values or the scaling, or it fixes a parameter that the data determine only weakly or only in combination with another \citep{wieland2021structural}. A structural repair adds, removes, replaces or reconnects components. A flat direction, along which several parameter values fit equally well, calls for a numerical repair, while a residual that stays systematic at every parameter setting calls for a structural one, unless the input itself is reconstructed wrongly. Telling the two apart is the central judgement of the task.

When to stop is also left to the agent. For each experiment $e$ and output $j$, the residual of the delivered model can be set against the stated noise,
\begin{equation}
  r_{e,j} \;=\; \frac{\operatorname{RMSE}_{e,j}(M, \hat{\theta})}{\sigma_j},
  \label{eq:ratio}
\end{equation}
where $\sigma_j$ is the noise level of output $j$ given in the task. A model with the right structure and well-fitted parameters, driven by the input as it was applied, reaches a ratio close to one, while a systematic error leaves it clearly above one, already on the experiments the model was fitted to. This reading presumes data that are a model plus noise, as on the four simulated systems. On a real device a deviation can be a local event, such as the bounce of the armature, and the magnet is therefore judged on features instead. Every quantity in this ratio is at hand in every round, and whether the agents use it is one of the things we observe. Section~\ref{sec:design} turns the ratio into the behaviour layer and adds the layers above it.

\section{Systems, Acceptance and Protocol}
\label{sec:design}

The five systems are ordered by how much is known about the answer at acceptance. The first three are textbook systems with a known answer, and each is built so that one question can be answered only from the data. In task~1, a spring--mass--damper, the description names no damper, so the dissipation has to be read from the traces. In task~2, a sled on a spring, it names no friction, and one force step stays below the breakaway force. In task~3, a DC motor with a flywheel, the armature inductance was chosen so that its effect stays near the noise: its electrical time constant of half a millisecond is a hundredth of the sampling interval, so the current has settled long before the next sample. Task~4 is the drive train of Section~\ref{sec:method}. Its architecture and the magnitudes of most of its parameters follow the electric drive of a published hybrid-vehicle study \citep{wallen2004hev}, while the composition and all values are our own. The battery characteristic shows only in the long experiment, and three pairs of parameters enter the outputs only in combination. For these four systems the reference is an assembly of library components that the agents never see, and the measurements are simulated from it with the stated noise added (Appendix~\ref{app:tasks}). Textbook physics cannot be kept out of an LLM and is part of what we study. Recall of public library examples is a different matter, so the references are not taken from such examples, and names and descriptions are varied. A risk remains, since only the parameter values and the inputs are certainly unseen.

The fifth system, task~5, has no reference model, since its data come from a real device rather than from a simulation. It is a flat-armature switching magnet whose static and dynamic behaviour was measured in the laboratory \citep{schmidt2019magneticforce,schmidt2020ironlosses}. On this task the client's side is taken by the builder of the device, who supplied its component data, wrote its briefing and rated the deliveries. No lumped network is expected to reproduce such a device in every detail. A bounce after an off-centre impact, for instance, can shift an edge in time without making the model any less valid. In place of the residual, the behaviour layer therefore uses a list of nine features with stated tolerances: three closing times, two release times, the plateau current and three peak closing speeds (Appendix~\ref{app:features}). The list was fixed after the first run of the base cell and before any further run. Residuals are still reported, against a scale of 33\,$\mu$m, 0.06\,m/s and 0.2\,A. This scale is the root-mean-square error that switching instants misplaced by 0.1\,ms produce over an experiment.

The magnet is handed over in four versions, with the same device and dynamic experiments but a different description or magnetisation curve. The base version is the text the builder first wrote, with the number of turns of the coil, the rate of the return spring, the outer dimensions of the yoke and a magnetisation curve measured up to a field strength of 1\,kA/m. The briefing is the text the builder would hand to an engineering office. It contains a cutaway drawing with one stated length, a magnetisation curve measured up to 60\,kA/m, which for technical purposes reaches saturation, and the indicators on which the model will be assessed. The briefing changes several things at once on purpose, because this is how a component is handed from one company to the next. Two controls separate the magnetisation curve from the rest. One keeps the briefing but replaces the drawing by a list of the same parts and figures, and the other keeps the base text and only replaces its curve by the one up to 60\,kA/m. The briefing was issued to four configurations: the smallest open-weight LLM and three of those that had met every feature at least once under the base version. The controls were run for the smallest LLM only, and as a harness check the curve-only version was also run under the other two harnesses. Appendix~\ref{app:prompt-frame} gives the framing instruction that every run received.

The acceptance reads the delivery $S$ of Equation~\ref{eq:solution} in layers. Each layer is graded on the delivered file and recorded separately, because a single verdict would hide which property failed. The first layer, $V_1$, is conformance. The model has to translate, its physics has to be built from components of $\mathcal{L}_d$, and it has to respect the rule $\mathcal{C}$. The compiler decides the first, and a reading of the file against the rule decides the other two. A class written by the agent passes when it only connects library components, with no block that integrates or differentiates in place of the physics, or implements a measured characteristic as a static relation. Agreement with the hidden reference structure is not required, since the same physics can be built in several valid ways. Every layer is evaluated on every delivery the evaluator can simulate, whatever the verdicts of the others, and a run counts as solved when it passes $V_1$ and $V_3$. The second layer, $V_2$, compares a fixed list of fitted quantities $\hat{\theta}_i$ for each task with reference values and passes when each lies within its band $B_{T,i}$. Where the experiments do not determine single parameters, the list uses combinations. A model in which a listed quantity has no counterpart leaves $V_2$ undecided. On the synthetic systems the band is a relative tolerance around the reference value (Appendix~\ref{app:tasks}). It is one per cent on task~1, because a vibration model is used to place resonances, and two to five per cent on the others. The wider bands apply where a value is known less well because of dry friction, temperature or a split that the data cannot resolve. For the deliberately flat inductance it is a factor of three. On the magnet the bands come from the stated and measured values of coil, spring and moving mass. The third layer, $V_3$, is behaviour. The evaluator recomputes the ratio $r_{e,j}$ of Equation~\ref{eq:ratio} from the file, with the input as it was applied in each experiment, and requires
\begin{equation}
  V_3(S) = 1
  \iff
  \max_{e \in \mathcal{D}_{\mathrm{fit}} \cup \mathcal{D}_{\mathrm{val}},\; j} r_{e,j} \le \sqrt{2}.
  \label{eq:gate3}
\end{equation}
The threshold is an engineering criterion. For zero-mean noise of variance $\sigma^2$ and a model error $b$ uncorrelated with it, the expected squared residual is $\sigma^2 + b^2$, so the gate accepts models whose systematic error does not exceed the noise. Moving the threshold anywhere between $1.3$ and $1.6$ changes the number of passing conformant synthetic runs only from 79 to 82. A run reaches the noise floor when it passes $V_3$, and it solves a synthetic task when it passes $V_1$ and $V_3$. On the magnet a run solves the task when it passes $V_1$ and meets every feature, and Section~\ref{sec:results} shows that this gate is lenient. The validation experiment is visible to the agent, so it is a check on the supplier's side. Five further experiments, which no run ever saw, test the synthetic deliveries after the fact. One has a trapezoidal input, and four come from a family of random steps, random ramps, chirps and sums of two sines that was frozen before any delivery was simulated on it. They are reported beside $V_3$ and do not count towards it.

These layers certify behaviour, but the same behaviour can come from different models. A delivered model consists of a structure $G$, its components and their connections, and parameters $\theta$. We write $[G]$ for the structure without sensors, sources, reference points and arithmetic on signals. Two structures are equivalent in mechanism when they produce the same outputs on the same parameters, and equivalent in assembly when they also use the same components. A constant friction torque and viscous damping are different mechanisms, whereas a combined spring--damper and a spring and a damper in parallel are one mechanism in two assemblies. The admissible set of a task is
\begin{equation}
  \mathcal{G}^{*}(T)
  =
  \bigl\{\, [G] \;:\; \text{there is a } \theta \text{ for which } M(G,\theta) \text{ passes } V_1 \text{ and } V_3 \,\bigr\},
  \label{eq:solutionset}
\end{equation}
and we count it both in assemblies and in mechanisms. A delivery belongs to it when it passes $V_1$ and $V_3$, so the layers above can still reject its members. The deliveries give a lower bound on the size of this set, while a benchmark that scores agreement with the hidden reference assumes that it has one element. How strongly each experiment responds to each parameter, one at a time, follows from the reference models (Appendix~\ref{app:resolution}). A one per cent error in the mass of task~1 moves the validation experiment to about thirteen times the noise, while on task~4 a damping error of five per cent stays at $1.06$ times the noise and passes.

\begin{figure}[t]
  \centering
  \includegraphics[width=\textwidth]{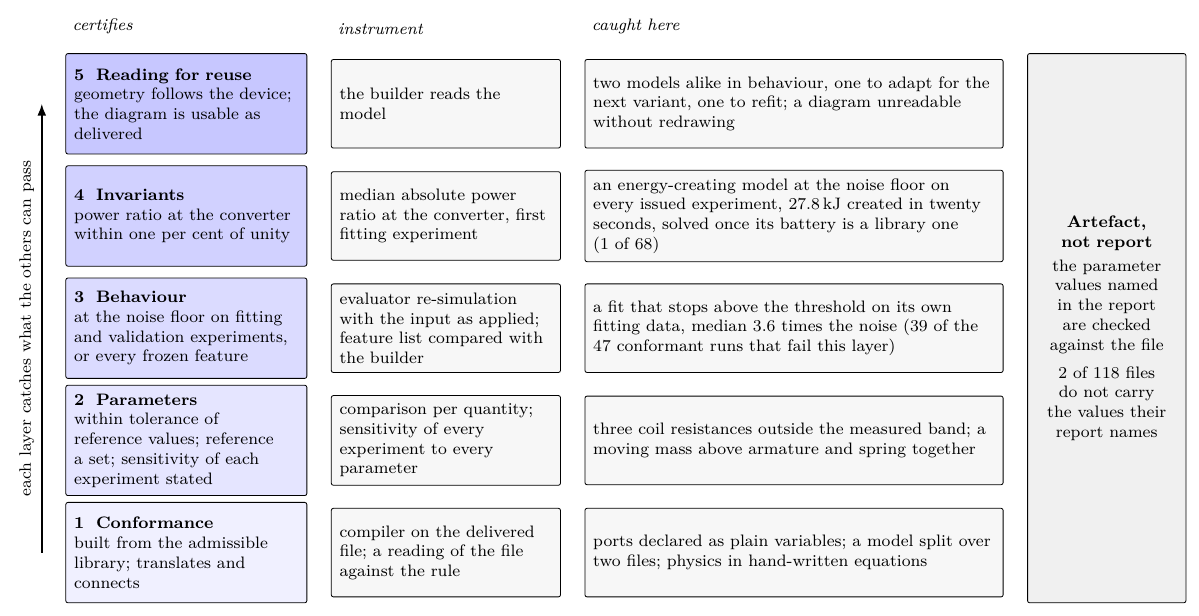}
  \caption{The acceptance scale: for each layer, what it certifies, the instrument, and a failure from this study that the layer catches. Each layer catches failures that the other layers can pass. All layers are graded independently; a run is solved when it passes conformance and behaviour. The reading was applied to seven magnet models; the check of the file against its report runs alongside the layers.}
  \label{fig:abnahme}
\end{figure}

Above behaviour the acceptance checks properties that cannot be seen in the behaviour (Figure~\ref{fig:abnahme}). The fourth layer, $V_4$, is a physical invariant. In general it is a power balance, in which the power entering a model equals the power leaving it plus the change of stored energy and the dissipation. On the two tasks with an electromechanical converter we check its simplest form at the converter, which neither stores nor dissipates energy. Every model is simulated on the first fitting experiment, and $V_4(S) = 0$ when the median absolute ratio of mechanical to electrical power at the converter differs from one by more than one per cent. In every delivered converter the ratio is the same at every sampled instant at which power flows, to within $10^{-15}$, so the median stands for all of them. Alongside the layers, the report is treated as a claim about the file. The fitted parameter values it names are compared with the values the file carries. The last layer is a reading. The builder read seven magnet models from six configurations, two of them from Opus~5, without knowing which LLM had delivered which. The reading asks whether the magnetic network is physically sound and whether its parameters follow the geometry closely enough to adapt the model to a variant instead of refitting it. It also asks whether the diagram can be worked with as delivered.

The suppliers are sixteen LLMs. Two Anthropic LLMs run under Claude Code, three OpenAI LLMs under Codex, and eleven open-weight LLMs from 28 billion to 2.8 trillion parameters under OpenCode. Since each LLM runs under one harness, we compare configurations. A configuration is an LLM with its harness, provider and settings. One configuration was also run under the other two harnesses. The open-weight LLMs below 35 billion parameters are included because an engineering office could run them on its own hardware, where the descriptions and measurements of a product often have to stay. Every configuration runs at the highest reasoning setting its provider offers. Every run gets one attempt, with no feedback and no second try, in a fresh process and working directory. Runs ended by a cause outside the configuration, such as an exhausted credit, were replaced, while every end caused by the agent or its harness counts. Each cell of task and configuration in the base matrix and under the briefing has at least three runs. Where the first three leave the verdict unclear, because they disagree on conformance or behaviour or all lie within ten per cent of the behaviour threshold, two more runs follow. Three to five runs show whether a configuration repeats a result, but they do not show how often it would. Three successes in three are compatible, at 95 per cent confidence, with a success rate as low as 0.29. A continuation rule ends an LLM's participation after three tasks without a run at the noise floor. Seven LLMs stopped without a run at the noise floor (identifiers in Appendix~\ref{app:serving}): Gemma~4 (31\,B), gpt-oss-120b, Mistral Medium~3.5 (128\,B), Muse Glimmer (30\,B), Nemotron~3 Super (120\,B), GLM-4.7-Flash (30\,B) and MiniMax~M3 (428\,B). The counts of Section~\ref{sec:results} cover the nine that continued. Because of these rules the number of runs differs between cells, so we compare cells and residuals rather than pooled rates. Network access is a declared tool, and no run searched for the published drive-train study or the publications on the magnet.

The task texts and measurements, conformance, the bands of the synthetic systems and the behaviour threshold were fixed before any run was graded. The magnet's bands for mass and coil resistance were fixed after its runs, from the measurement record. The invariant, the check of the file against its report, the test experiments, the variants and the black boxes came later, and each of these checks was applied to every delivery to which it applies. The briefing, its controls and the harness check followed the base results.\repoline

\section{Results}
\label{sec:results}

\begin{figure}[t]
  \centering
  \includegraphics[width=\textwidth]{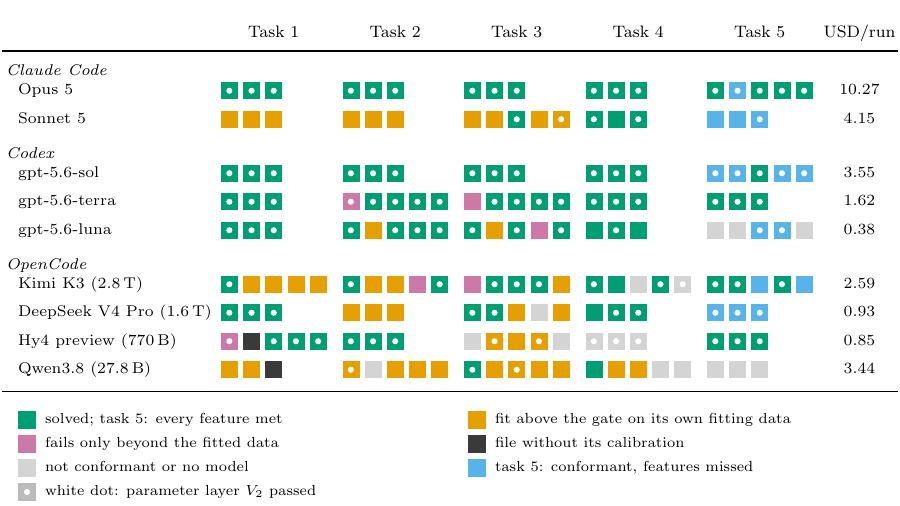}
  \caption{The runs of the base matrix for the nine LLMs that reach the noise floor at least once, one square per run in the order the runs were made. On tasks~1 to~4 the colour gives the group into which the precedence rule of this section sorts the run, and on task~5 whether the run delivered a conformant model and met every feature of the frozen list. A white dot marks a run that passes the parameter layer $V_2$; runs without a dot fail it or offer nothing to compare with the reference. Right: median cost of a run that delivered a conformant model on tasks~1 to~4, in US dollars at list prices; the smallest LLM is cheaper per token, but its runs are several times longer; each run is a single attempt without feedback. Sizes are the total parameters of the open-weight LLMs.}
  \label{fig:matrix}
\end{figure}

The base matrix of the nine LLMs that reached the noise floor at least once comprises 138 graded runs on the four synthetic tasks, in 36 cells of task and configuration with at least three independent runs each, and 35 on the magnet in nine cells (Figure~\ref{fig:matrix}).

Two of the sixteen configurations, Opus~5 under Claude Code and gpt-5.6-sol under Codex, solve all four synthetic tasks in every repetition. Five more solve at least one task in every repetition and are split on others. The smallest open-weight LLM, Qwen3.8 with 27.8 billion parameters, reaches the noise floor in single runs but in no complete cell, and the seven that stopped never reach it. Their failures are recognisable. Some deliver models that translate but stay far above the noise, or leave the start values in the file while the report names a fit. Others answer with a single sentence and no tool call, invent class names the library does not contain, or write their files outside the working directory. Conformance is not the hurdle for most of the nine: five clear it in every run, DeepSeek, Kimi~K3 and Qwen3.8 miss it in one to three runs, and Hy4 in five of sixteen. The reading of the files changes this picture on task~4. Six deliveries there pass behaviour but break the library rule, all at the battery: three runs of Hy4 count the charge with an integrating block, and two of Kimi~K3 and one of Qwen3.8 with an integrator or a derivative of their own. None of the six reports flags a breach. Several argue that the construction is admissible, one claims that no library component provides a charge-dependent source although the library has two, and one, the model of the introduction, states that only admissible library blocks are used. All six come from open-weight LLMs under OpenCode, where in our design the model family and the harness coincide. A client who read only the curves and the report would have accepted all six. Apart from that, the smallest LLM fails in the same way as the mid-ranked LLMs, in the modelling. It lands in a second optimum on task~1, fits a flat direction on task~3 and chooses the alternative dissipation mechanism on task~4, which passes the behaviour layer. A single run does not show which configuration delivers. Of the 36 synthetic cells, sixteen have every run solved, thirteen are split and seven contain no solved run. Where the calibration can be compared, it holds: 74 of the 80 solved runs pass the parameter layer, and the other six, all on task~4, carry no counterpart to a listed quantity, which leaves the layer undecided. Thirteen of the 74 round their fitted values, and since the reference values are round numbers, they land exactly on them. In one split cell the three runs stand at $1.03$, $1.14$ and $1.93$ times the noise, and the protocol extends such cells to five runs.

Reaching the noise floor does not single out a model. Counting the distinct assemblies among the runs at the noise floor, in the sense of Equation~\ref{eq:solutionset}, gives at least two on task~1, five on task~2, two on task~3 and eight on task~4. The library settles some of these differences by itself, since one library component can be defined as the composition of others. On task~1, for instance, a \texttt{SpringDamper} is a spring and a damper in parallel. Where the components carry different equations, we rebuilt every counted assembly with one common parameter set and simulated the variants side by side (Appendix~\ref{app:assemblies}). On tasks~2 and~3 the assemblies of a task, set to one common parameter set, behave as one physical model and differ only in which component carries which effect. On task~4 two mechanisms survive: the viscous dissipation of the reference, and a constant friction torque that the acceptance cannot reject. Setting it to $17.36$\,N\,m and changing nothing else keeps the deviation from the reference below the threshold in every experiment, at most $1.1$ times the noise in the current. The three solved runs that state it at the output shaft come from three vendors and carry $17.36$ to $17.47$\,N\,m. The two cannot be told apart because after the run-up the output speed stays within 15 per cent of its mean in every experiment. The viscous torque, 14 to 17\,N\,m at the mean speeds, therefore varies by less than 3\,N\,m, while the brake applies 50 to 600\,N\,m. The other differences on task~4 are arrangements that no output reveals, such as whether rotor and load are one inertia or two. Outside this count, since it fails conformance, one delivery rebuilds the converter with two separately fitted constants, and only $V_4$ of the layers sees that error. Set to a common parameter set, the assemblies at the noise floor thus realise one dissipation mechanism on tasks~1 to~3 and two on task~4, while the library decides how many arrangements a mechanism can take. A benchmark that reports the second number as structural diversity measures its own component library. The data themselves are not neutral. Fitted alike, viscous damping and Coulomb friction both stay within $1.03$ times the noise on every issued experiment of task~4, yet differ by about 300 in $\chi^2$ over some $14\,000$ residuals (Appendix~\ref{app:resolution}). Statistically the data favour one mechanism, but at the accuracy that an acceptance can agree on, both are admitted. No brake profile we simulated changes this: the fitted Coulomb model stays within $1.1$ times the noise up to twice the largest issued load, and only a load that pulls the drive close to standstill, at about four times the current, lifts it above the gate.

Passing the behaviour layer also does not exclude a model that violates physics. The model of the introduction is a run on task~4 by the smallest open-weight LLM. Its residual is at most $1.03$ times the noise on every issued experiment, and its torque constant matches the reference within a per cent. It fails conformance, because its battery integrates the state of charge in an equation of its own, but that is not where the physics breaks. Its back-EMF constant, however, is $k_e = 0.614$ against $k_t = 1.158$, because the run rebuilt the coupling from a voltage source, a torque source and two gains (Figure~\ref{fig:converter}) and fitted the two constants separately. Simulated on the first fitting experiment, it gives
\begin{equation}
  \frac{P_{\mathrm{mech}}(t)}{P_{\mathrm{el}}(t)} = 1.8852
  \quad\text{throughout},
  \qquad
  31.4\,\text{kJ in},\; 59.2\,\text{kJ out in twenty seconds},
  \label{eq:powerratio}
\end{equation}
so $V_4(S) = 0$ while $V_3$ holds, and over the experiment the model creates $27.8$\,kJ, 89 per cent of the electrical energy it draws. The coupling itself is built from admitted blocks. With its battery rebuilt from a library capacitor, which leaves the equations unchanged, the file passes $V_1$, $V_3$ and all five test experiments, counts as solved and still fails $V_4$. The measurements cannot reveal this, because the parameters are not identifiable. With the two constants split, the armature equation $L\,\dot{i} = U - R\,i - k_e\,\omega$ keeps its solution when the battery voltage $U$, the resistances $R$, the inductance $L$ and $k_e$ are scaled by one factor. The torque $k_t\,i$ does not change, so current and speed stay exactly the same, and only the power ratio $k_t/k_e$ moves. The identity $k_e = k_t$ is what pins this factor. The run has landed elsewhere in this family, with its back-EMF constant, battery voltage, resistance and inductance all about $1.87$ times below the reference. $V_2$ stays undecided, since the model offers two constants where the reference has one. Mapped onto the reference, it would fail on resistance and inductance. On a real device such a check needs independent values for the electrical side, while the power balance needs none. The signal path as such is not the defect. A task-3 run of the same LLM rebuilds the coupling in the same way but declares the constant once, and it passes every layer. Of the 68 models with a converter that the evaluator can simulate, 64 take the coupling from the library, where the balance holds by construction. Four rebuild it, all from this LLM, and only in this one do the two constants differ. $V_4$ holds for 67 models and fails for this one. A census of all 162 models of the base matrix that the evaluator can simulate finds no other place where the balance breaks by a noticeable amount (Appendix~\ref{app:assemblies}).

The report is the next claim the acceptance reads against the file. For 118 of the 127 conformant synthetic runs, the fitted values named in the report can be matched to the reference. In two of them the file carries other values, the start values $\theta_0$ instead of the result of the calibration, and both fail the behaviour layer. The check therefore catches no model that the layers passed, but it shows that a report cannot stand in for its file. In one of them, a run of Hy4 on task~1, the report lists a fit within a tenth of a per cent of every reference value, while the file carries the start values. The evaluator, scoring the file, finds a residual 750 times the noise. The omission is easy to make. The agents fit outside the file, by overriding the parameters of the compiled model, by setting them through a scripting interface or by integrating their own copy of the equations, and the file changes only if the agent writes the result back.

The behaviour layer can pass a black box. As an illustration, not as the best a system identifier could build, we identified two black boxes from the same fitting experiments: a linear output-error state-space model whose order is chosen by leaving one fitting experiment out at a time, and the same model with a neural correction, chosen in the same way (Table~\ref{tab:baseline}, Appendix~\ref{app:tasks}). On task~1 the linear model of second order passes the layer at $1.02$ on the fitting and $1.15$ on the validation experiment. It also passes all five test experiments, and only conformance rejects it, since state-space blocks are not admitted. Its coefficients even reproduce the physical parameters to about a tenth of a per cent, but they belong to no component. On the sled, where three force levels leave the friction law open, neither of the two black boxes we identified passed the validation experiment. Once the device changes, the difference between the two kinds of model is plain. We built variants of each device in which one parameter is raised by half, the mass or the spring rate on tasks~1 and~2 and the inertia or the armature resistance on task~3. The same parameter was raised in the delivered models. All 39 delivered models of tasks~1 and~3 that pass the layer stay within the gate on every experiment of the variants. On the sled, only 6 of 19 stay within it, and eleven of the others miss it narrowly on a single experiment. Every black box, left as identified, misses every variant, by a factor of about a hundred or more. Mapped back to mass, spring and damper, the linear model of task~1 meets both variants, at $1.13$ and $1.07$ times the noise, as the delivered models do. But that mapping is the physical structure, and it exists one to one only because the black box has the order of the device. Within the issued range, the five test experiments confirm 72 of the 80 solved runs. The eight that fail do so on the two quantities the issued experiments resolve least, the ratio of static to sliding friction on task~2 and the inductance on task~3. Beyond the issued range the delivered models part ways, most clearly where the data left parts of them undetermined. On task~3 the worst ratio at three times the input amplitude grows with the deviation of the delivered inductance, from at most $1.35$ at the reference value to $6.3$ at half of it. Passing the acceptance thus certifies a model only as far as its experiments resolve it.

\begin{table}[t]
\centering
\caption{Worst ratio of residual to noise per group of experiments. \emph{Delivered}: range over the runs that pass the behaviour layer (19, 19 and 20). \emph{Linear}: output-error state-space model. \emph{Network}: the linear model with a neural correction, chosen like it by leaving one fitting experiment out, range over three seeds; on task~1 the selection returns the linear model. The columns $\times 2$ and $\times 3$ scale the inputs of the test trapezoid, chirp and random steps by two and three, and that of the first fitting experiment by three. Variants raise one parameter by half, in the device and in the delivered model, and the column gives the range over both variants. The networks are simulated in Python with the input as applied, all other values with the test bench of the acceptance. Task~4 has neither variants nor black boxes.}
\label{tab:baseline}
\small
\setlength{\tabcolsep}{4pt}
\begin{tabular}{@{}llrrrrrr@{}}
\toprule
Task & Model & Fit & Validation & Test & $\times 2$ & $\times 3$ & Variants \\
\midrule
1 & delivered & 1.02 & 1.02--1.24 & 1.05--1.06 & 1.04--1.18 & 1.06--1.33 & 1.05--1.11 \\
  & linear & 1.02 & 1.15 & 1.06 & 1.18 & 1.29 & 274--387 \\
\addlinespace
2 & delivered & 1.01 & 1.03--1.23 & 1.05--31 & 1.03--11 & 1.01--1035 & 1.01--4.9 \\
  & linear & 56 & 406 & 293 & 724 & 1154 & 392--1073 \\
  & network & 12--22 & 178--216 & 222--324 & 618--758 & 1023--1199 & 473--1474 \\
\addlinespace
3 & delivered & 1.06--1.07 & 1.02--1.27 & 1.05--2.26 & 1.05--4.21 & 1.03--6.28 & 1.06--1.17 \\
  & linear, chosen order & 1.08 & 1.81 & 2.48 & 4.68 & 6.99 & 206--207 \\
  & linear, order 4 & 1.06 & 1.17 & 1.39 & 2.19 & 3.04 & 206--207 \\
\bottomrule
\end{tabular}
\end{table}

\begin{figure}[t]
  \centering
  \includegraphics[width=\textwidth]{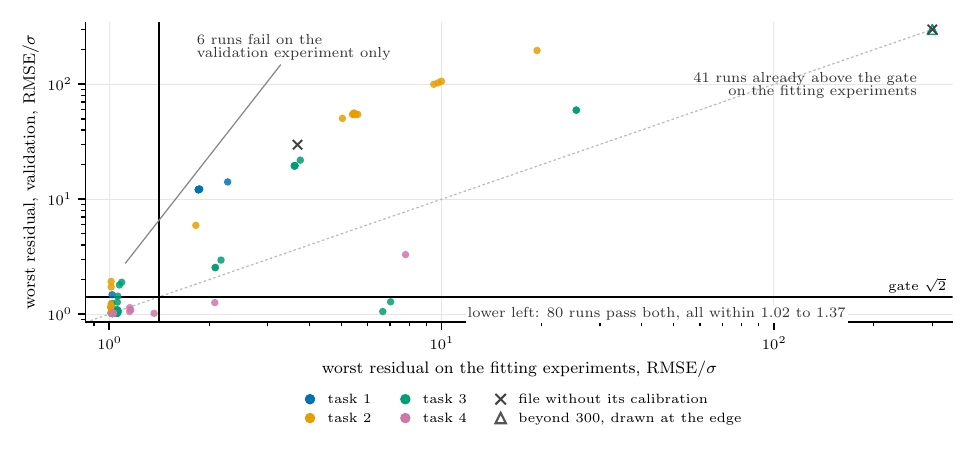}
  \caption{Every conformant graded run of the nine LLMs on the first four tasks: worst residual on the fitting experiments against worst residual on the validation experiment, in units of the disclosed noise. The gate of Equation~\ref{eq:gate3} divides the plane, and the dashed diagonal marks equal residuals on both. Colour gives the task; crosses mark the two files, both on task~1, that do not carry their calibration. Points beyond 300 sit at the edge. The eleven runs that are not conformant are not shown.}
  \label{fig:residuals}
\end{figure}

The runs that fail the behaviour layer mostly fail before the validation experiment. Of the 47 conformant synthetic runs that fail it, 41 already exceed the threshold on the experiments they were fitted to. Only six meet the fitting data and fail on validation, at $1.43$ to $1.93$ (Figure~\ref{fig:residuals}). Two of the 41 are the files without their calibration. The remaining 39 lie at a median of 3.6 times the noise on their own fitting data, with an interquartile range of about two to six. We sort the 138 graded synthetic runs by a precedence rule into five groups. There are eleven runs that are not conformant, five of them without a model the evaluator can simulate, two files without their calibration and 80 solved runs. The fourth group holds these 39 fits delivered above the threshold on their own fitting data, two thirds of the failures, and the fifth holds the six runs that fail only beyond the fitted data. Raising the threshold to three times the noise leaves the fourth group the largest group of failures, with 25 runs against 10, and the fifth group overtakes it only at about five times the noise. Weighting every cell equally leaves it the largest as well, at 60 per cent of the failures. Each of the 39 comes from an LLM that reaches the noise floor on another task. The dominant failure is therefore a fit that stops while its own data still show a systematic residual. The task text did not contain the threshold, but it contained what was needed to notice this, since the noise level is stated, the RMSE is requested, and the comparison takes one division.

What the agents did with this evidence can be read in their own reports. We coded the 39 runs by rules fixed before the counting. Two LLM coders worked independently and agreed on 35 runs (Cohen's $\kappa = 0.83$), and we resolved the other four against the run files. One run ends without declaring the delivery complete, cut off at the output limit in the middle of an action. Each of the other 38 reports an RMSE for every fitting experiment, so the comparison with the stated noise was at hand, and in 37 of them the agent's own numbers already exceed the threshold. In 23 of them the report names a residual above the noise or a limitation of the structure, and the agent delivers all the same. The deviation is often explained away as a property of the noise or of the admissible library. In 10 the report describes a fit as close to the noise although its own numbers place it above the threshold. In four the report gives the numbers without judging them, and in one it gives errors below the threshold, while the evaluator's simulation of the same file exceeds it. If only residuals on the fitting data count as named, the 23 become 20 and the four become seven. The loop thus breaks in two ways. Some agents miss or misjudge the comparison, and at least as many make it and deliver anyway.

Some of the 39 land at the same wrong values across model families. On task~1 eight runs from three LLMs place the mass $1.04$ to $1.06$ per cent low, at $1.86$ on their fitting data and $12.1$ to $12.3$ on the validation experiment. On task~2 eight runs from four vendors miss the mass by $2.47$ to $2.64$ per cent. Two of them agree to four significant figures, although one comes from a dense LLM of 27.8 billion parameters and the other from a mixture of experts of $1.6$ trillion. They are not optima of the physical system. Driven by the input as it was applied, the same structure fits at the reference values. All 16 runs of the two groups feed the sampled input into their simulation by linear interpolation. Linear interpolation turns each edge of the pulses and steps into a ramp over one sampling interval and so delays it. No parameter of a correct structure can make up for that: under this input even the reference values come no closer than $2.1$ and $6.2$ times the noise on the first fitting experiment. With the input reconstructed in this way, the best fit of the same structure lies close to the values these runs share (Appendix~\ref{app:resolution}). The recurring error points to the objective the agents construct rather than to the LLM. On tasks~1 and~2, the runs that solve differ from those that deliver above the threshold in this reconstruction of the input. Two LLM coders who did not know the verdicts read the 38 solved runs and the 22 fits delivered above the threshold on these two tasks. None of the 38 solved runs feeds the samples in by plain linear interpolation, against 17 of the 22 others. Checking the residuals for structure, by contrast, does not distinguish the solved runs: 16 of the 38 did so, against 19 of the 22 others. The handover allowed a reconstruction good enough to pass, since the solved runs worked from the same samples and the same text.

A follow-up asks what a stated acceptance criterion changes. For two configurations we repeated the three cells in which they had most often delivered above the threshold: Sonnet~5 on tasks~1 and~2 and Qwen3.8 on task~2. The repetition ran with concurrent controls under the original text, and only the task text changed between the arms. One version adds the criterion, a ratio of at most $\sqrt{2}$ on every fitting and validation experiment, and a second adds the instruction to compare, to repair above it and to say so if delivering above it (Appendix~\ref{app:tasks}). Cells, controls and analysis were fixed before the first run. Under the original text all nine control runs again deliver above the threshold, and four of their reports describe the fit as close to the noise. With the criterion stated, 23 of the 24 runs compare their errors with it, and the other ends without a report. None describes a fit above it as close to the noise. Under the original text, under the criterion alone and under the criterion with the instruction, Sonnet~5 solves task~1 in none of three, three of five and four of five runs. On task~2 it solves none of three, none of three and one of five, and Qwen3.8 solves none of three, two of three and three of three. Together, 13 of the 24 runs with a stated criterion solve the task, against none of the nine controls, and among the first three runs of each arm and cell, 9 of 18 against none of 9. The outcome follows the reconstruction of the input, since every run that solves reconstructs it by a method other than linear interpolation. On task~2 six runs of Sonnet~5 stay above the threshold on their fitting data, four of them at the recurring optimum, and every one of their reports says so. The cells were chosen where the configurations had failed, so the follow-up shows what the criterion changes there, not what it does for configurations that solve without it.

On the magnet the acceptance has no reference and no reachable noise floor, and its layers behave differently. Of the 35 graded base runs, the residual gate of Equation~\ref{eq:gate3}, applied to the stated scale for comparison, passes one, at $1.4139$, while the frozen feature list passes fourteen (Table~\ref{tab:task05-cells}). Of the eighteen runs under the briefing, the residual gate passes none. The two instruments measure different things. The best run reproduces plateaus and switching times and departs in transients a few milliseconds long, and a residual over a trace with sharp edges is dominated by where the edges lie. The features ask whether each switching event happens at the right time and with the right shape, which is what a switching magnet is for. The builder rated the standard charts of 28 of the 29 conformant base runs, on 30 criteria and without seeing features, verdicts or the delivering LLM. The mean rating correlates with the number of missed features at a Spearman coefficient of $-0.80$ (95 per cent interval $-0.87$ to $-0.57$, resampled over configurations). As a gate the list is lenient. It passes five of the seven runs that the builder rates `not good' overall, though none of the ten rated worst. The parameter layer separates little here, and 25 of 29 conformant runs pass it. The quantities with a reference value sit at the terminals of the device: the turns of the coil and the rate of the return spring, stated in the brief, the moving mass of armature and spring, bounded by weighing, and the coil resistance, read from the current plateaus of the issued experiments. The pole radii, gap areas and leakage paths decide whether a model is a model of this device, and none of them has a measured value in the package.

\begin{table}[t]
\centering
\caption{The magnet by configuration: graded runs, features missed per run against the frozen list, and the range of the worst-output residual in units of the stated scale of Section~\ref{sec:design}. Shown are the base handover, the builder's briefing for the four configurations it was issued to, and the smallest configuration under the two controls and the harness check. The briefing cell of Kimi~K3 has three runs although they disagree: its diverging run misses the same two features as that of gpt-5.6-sol, so further runs would have enlarged the sample without settling the question. The OpenCode row of the controls pools five runs at the harness defaults with three in which the context length was declared (Appendix~\ref{app:serving}).}
\label{tab:task05-cells}
\small
\begin{tabular}{@{}lccl@{}}
\toprule
Configuration & Runs & Features missed & Worst residual \\
\midrule
\multicolumn{4}{@{}l}{\textit{Base handover}} \\
gpt-5.6-sol   & 5 & 0, 1, 1, 5, 6 & 1.6 -- 2.2 \\
Opus~5                 & 5 & 0, 0, 0, 0, 1 & 1.7 -- 3.5 \\
gpt-5.6-terra & 3 & 0, 0, 0       & 1.5 -- 4.3 \\
Sonnet~5               & 3 & 4, 6, 6       & 3.9 -- 11.0 \\
DeepSeek V4 Pro (1.6\,T) & 3 & 2, 4, 6       & 1.9 -- 6.1 \\
Hy4 preview (770\,B)   & 3 & 0, 0, 0       & 1.7 -- 2.3 \\
Kimi~K3 (2.8\,T)       & 5 & 0, 0, 0, 4, 4 & 1.41 -- 3.4 \\
gpt-5.6-luna & 5 & 4, 5; three fail $V_1$ & 2.0 -- 9.8 \\
Qwen3.8 (27.8\,B) & 3 & \multicolumn{2}{l}{no conformant model delivered} \\
\addlinespace
\multicolumn{4}{@{}l}{\textit{Builder's briefing with drawing}} \\
gpt-5.6-sol   & 5 & 0, 0, 2, 0, 0        & 1.42 -- 2.22 \\
Opus~5                 & 5 & 0, 0, 1, 0, 1        & 1.89 -- 2.49 \\
Kimi~K3                & 3 & 0, 0, 2              & 1.50 -- 2.35 \\
Qwen3.8                & 5 & 2, 2, 3, 4, 5        & 3.27 -- 7.47 \\
\addlinespace
\multicolumn{4}{@{}l}{\textit{Qwen3.8 under the controls and the harness check}} \\
briefing without drawing, OpenCode & 3 & 2, 6, 5 & 3.27 -- 9.02 \\
base text, curve to 60\,kA/m, OpenCode & 8 & 1, 2, 1, 6; four deliver nothing & 2.78 -- 11.60 \\
\quad same, Codex                   & 3 & 6, 2; one delivers nothing & 2.33 -- 11.40 \\
\quad same, Claude Code             & 3 & 1, 1, 1 & 3.82 -- 9.56 \\
\bottomrule
\end{tabular}
\end{table}

\begin{figure}[t]
  \centering
  \includegraphics[width=\textwidth]{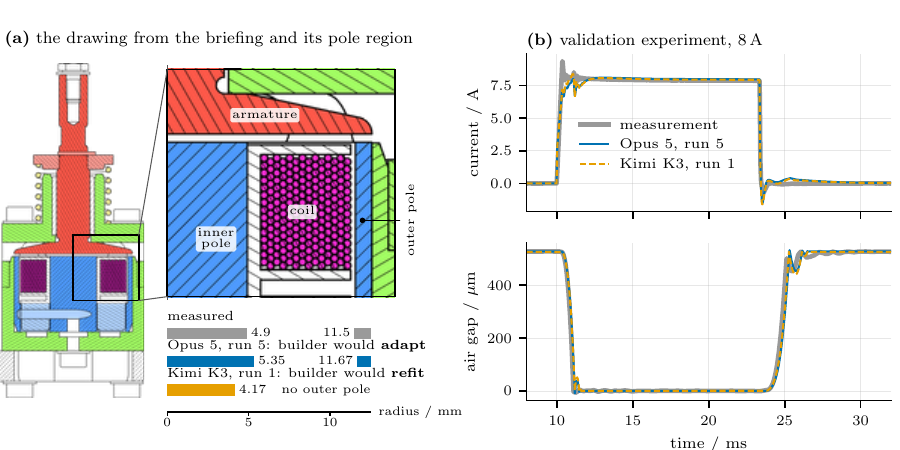}
  \caption{The decisive pair on the magnet. (a)~The builder's cutaway drawing from the briefing, scaled by its one stated dimension, with the pole region enlarged; below it the pole faces as measured and as the two delivered models carry them, and the builder's verdict on reuse (Table~\ref{tab:expert-models}). (b)~The validation experiment, current-controlled at 8\,A: the measurement and both delivered models, simulated as the evaluator does with the measured coil voltage as input. By the builder's rating of their behaviour the two models cannot be told apart; the builder would adapt one of them to a new geometry and refit the other.}
  \label{fig:pair}
\end{figure}

Only a reading can check these quantities, and the builder read seven of the models (Figure~\ref{fig:pair}). Both models of the decisive pair are in the builder's top tier. Opus~5 run~5 and Kimi~K3 run~1 cannot be told apart by behaviour. Both meet every feature, their residuals on the comparison scale are $1.73$ and $1.68$, 29 of 30 criteria are good or reasonable for each, and the builder places the difference within the uncertainty margin. In structure the two differ clearly. The Opus model assembles yoke, poles and armature from flux tubes with geometric parameters. It fits the pole radii within a tenth of the measured values and the coil resistance inside its band, and the builder judges that it can be adapted to a new geometry. The Kimi model fits lumped lengths and areas that follow no dimension of the device and carries the residual air gap twice, and the builder would refit it (Table~\ref{tab:expert-models}). Within one configuration the same contrast appears in the dimensions rather than in the network. Opus~5 run~1, the run with the highest mean rating, is element for element the network of run~5, with dimensions fitted where run~5 fixed them. Under the briefing with the drawing the dimensions stop being fitted, and four of the five Opus runs carry the pole radius within one per cent of the measured value, a dimension that no task text states and that can only be read off the scaled drawing.

For the smallest open-weight LLM the handover made the largest difference we saw. Under the base text none of its runs delivers a model the evaluator can simulate, for three different reasons: no output connector, ports declared as plain variables, and a needed class in a second file. Under the briefing every run delivers, with the drawing and without it, so for this LLM delivery does not depend on the drawing. The curve up to 60\,kA/m, with the text otherwise unchanged, lets nine of fourteen runs across the three harnesses deliver, and five of those nine miss a single feature, against two to six under the briefing with and without the drawing. Every harness delivers in at least half of its runs under the curve-only handover, so none of the three is what blocks delivery.

\section{Discussion}
\label{sec:discussion}

Passing the behaviour layer certifies a model as far as its experiments resolve it, and even there the structure stays open, by an amount that depends on the accuracy agreed. On the drive train the data favour viscous damping, yet an acceptance at a tolerance passes both mechanisms. The admissible set is thus a property of the agreed accuracy as much as of the data. Some errors stay hidden at any accuracy. A converter whose one constant has become two reproduces the measurements just as well, and more measurements of the same kind cannot expose it. This is why the acceptance has layers above behaviour, the invariant at the cost of a simulation and the reading at the cost of an expert's time. None of this requires a reference. On the magnet, where the residual gate passes one run of 35, the behaviour layer uses a frozen list of features instead, while the construction of the acceptance stays the same.

For machine learning the methodological point is the reference. A benchmark that scores agreement with one reference structure assumes that the answer is unique. When noisy data have to settle what the description leaves open, the answer is a set. It is the structural counterpart of the Rashomon set of models with near-optimal accuracy \citep{fisher2019all}, and its size depends on the accuracy one agrees on. A score against one reference then also reflects the component library, which multiplies the arrangements in which each mechanism can be built. An acceptance that checks membership in the admissible set, as ours does, needs no reference structure and avoids this bias, while the library and the conformance rule still decide what the set admits. The argument does not depend on Modelica. It applies wherever a handover leaves the model underspecified \citep{damour2022underspecification} and the delivered model is meant to be read, not only run.

The gain from a decomposable model shows when the device changes. We required library components because a model whose parameters are properties of the device can be adapted to a variant by changing them. Within the issued range this advantage can be invisible, since on task~1 a linear black box passes the behaviour layer and the test experiments. It appears on a variant of the device: on the two linear tasks every delivery that passed the behaviour layer stays within the gate, and on the sled, whose friction law the data left open, only 6 of 19 stay within it. The black boxes, left as identified, miss every variant by a factor of about a hundred or more, and adapting one would first need the mapping to physical parameters that the component model carries by construction. On the magnet no variant was measured, so the property has to be read rather than simulated. The reading separates two models that behave alike: the builder would adapt one to a new geometry and refit the other. The client can also shape the property, as the briefing with its drawing did for Opus~5. In this sense interpretability can be tested where a variant exists, judged by an expert's reading where none exists and influenced through the briefing, and in each case on the delivered artefact.

The failure we met most often, fits delivered above the threshold on their own fitting data, sits where the acceptance criterion was left open. Part of this class goes back to the input, which the handover gave as samples, as a measurement does, so that the agents had to reconstruct it. The finding does not depend on that cause, because whatever produced the residual, the agents had the numbers that show it and delivered all the same. The base tasks stated the noise and asked for the errors, but not how close was close enough, so the requirement was not yet verifiable in the sense of requirements engineering \citep{iso2018requirements}. That is the normal state of a first handover. Requirements are refined in rounds between client and supplier, and an experienced engineer fills the gaps with a sense of what the client will accept. The follow-up shows what one round of refinement does. A stated criterion acts like a failing test. In the follow-up the agents compared, reported a shortfall as such, and improved where they then reconstructed the input by a method other than linear interpolation. The criterion did not supply that change. Where it was missing, the agents named the residual and delivered. The criterion thus fixes the comparison but not the diagnosis. A harness that makes the agent account for each obligation against evidence points in the same direction \citep{wang2026pufibara}.

The claims have limits that the design makes explicit. For the layers above behaviour, single cases show that they are needed. The invariant catches one of 68 models with a converter, and the reading covers seven models rated by one expert. These cases show that such failures occur and which instrument sees them, but not how often they occur across devices or readers. How far the findings carry depends on the systems. The first four are close to textbook examples, so recall from pretraining is part of what is studied and cannot be separated from reasoning. Their data are simulated, and only the magnet is measured. All five are lumped component models in the mechanical, electrical and magnetic domains. Every handover came with recorded data, and handovers with partial signals or vague descriptions, where requirements would have to be reconstructed rather than read, were not part of the study. Thermal and fluid systems were not tested, and nothing is claimed about systems whose physics resides in tabulated data. The protocol sets further bounds. The validation experiment is visible to the supplier and therefore no test in the strict sense \citep{dwork2015reusable}. The report of the magnet example in the supplement takes the coil resistance from the current plateaus of all three experiments, the validation experiment among them, and states that this experiment was never used in a fit. On the four synthetic tasks, five test experiments supply such a test after the fact. The invariant is checked at the converter, with a census of the other models rather than a balance over every model. Each variant raises a single parameter, and reuse on the magnet was judged by reading. Every run was a single attempt without feedback, whereas a client would answer a delivery that misses the acceptance. The results therefore describe what a supplier achieves on its own. And the layers are acceptance criteria for procurement, not a safety certification.

\section{Conclusion}
\label{sec:conclusion}

We set out to see whether LLM agents can do this work at all, and they can. On the four synthetic tasks they turn a description and measurements into system models that pass conformance and behaviour, two configurations in every repetition. What passing certifies is less than one would hope. It certifies that the model reproduces the measurements within the range they resolve. It does not certify the structure, since the measurements of the drive train admit two dissipation mechanisms at the agreed accuracy. It does not certify the physics, since a model that creates energy reproduces every measurement. Nor does it certify that the model can be adapted: of two magnet models that behave alike, the builder would adapt one and refit the other. These gaps call for different answers. The first lies in the handover, not in a model, and an acceptance can only check that a delivery belongs to the set of structures the handover admits. A benchmark that scores agreement with one reference assumes that this set has one element, and its score then depends partly on the component library. The other two can be checked at a price: a power balance at the converter catches the energy-creating model in one simulation, and an expert's reading separates the magnet pair. Alongside these layers the file is checked against its report, because a report can name values its file does not carry, and declare a model admissible that breaks the rule of the task.

What we did not know at the start was whether an agent would make the old mistake of taking a good fit for a valid model. It does, and not for lack of numbers. The largest group of failures are fits delivered while the agent's own fitting data still lie above the threshold, although the task stated the noise level. All of these runs come from LLMs that reach the noise floor on other tasks, and in about half of them the report names the deviation. What the task left open was how close is close enough. Once a criterion was stated, it made the agents compare and say where they fell short, and the deliveries improved where the agents changed how they reconstructed the sampled input. On the two tasks with recurring optima, the original reconstruction comes close to those optima. Where we did change the handover, the deliveries followed: given the builder's drawing, the configuration whose models the builder read in most detail takes dimensions from it instead of fitting them, and given an acceptance criterion, the agents compared their fits with it. Interpretability is thus a property of the delivered model that the client can influence.

Every run was a single attempt without feedback, as if the client could only take or reject the first delivery. The smallest LLM we tested makes the next step worth taking. It is small enough to run where the data have to stay, and on its own it reaches the noise floor of the synthetic tasks only in single runs. Under the builder's briefing, however, it delivered a gradable model of the real device in every run, and such a delivery is something a client can answer. Whether a few rounds of answers turn deliveries into accepted models is the experiment we want to run next. If they do, the agent becomes a supplier that the client can question, and accepting its model becomes what it is with an engineering office, the end of a conversation rather than a single verdict.

\bibliographystyle{unsrtnat}
\bibliography{references}

\appendix
\section{The Synthetic Tasks}
\label{app:tasks}

Table~\ref{tab:tasks} gives what is needed to rebuild the four synthetic tasks. Each reference is assembled from components of the Modelica Standard Library, except that task~4 integrates its state of charge by one equation. The rule does not admit such an equation in a delivery, but the reference is a device and not a delivery. Every experiment is simulated in OpenModelica from rest and sampled on a uniform grid, and Gaussian noise of the stated standard deviation is added from a fixed seed. The five test experiments are built the same way. The trapezoid runs twice within the amplitude range of the issued experiments of the task. The other four are drawn from seed 20261003: six random steps, eight ramp points, a linear chirp across the frequency band of the issued experiments and a sum of two sines at random frequencies and phases in that band, all within the issued range. The task text gives the agents the outputs, the noise levels, the kind of input in each experiment, the sampling and the admissible part of the library, but not the reference or its values, the acceptance threshold, the test experiments or the times at which an input switches. It asks for a single file \texttt{model.mo} with the named input and outputs, the fitted values and the errors on every experiment. For the follow-up of Section~\ref{sec:results} the texts of tasks~1 and~2 gain a section before the required output: ``The model is accepted if, for every measured output, its RMSE on each fitting experiment and on the validation experiment is at most $\sqrt{2} \approx 1.414$ times the noise standard deviation stated above for that output.'' The second version adds: ``Before you deliver, compare every RMSE with this threshold. If any lies above it, repair the model, numerically or structurally, and compare again. If you deliver a model that still lies above the threshold, say so in the report and explain why.''

\begin{table}[tp]
\centering
\footnotesize
\setlength{\tabcolsep}{3pt}
\caption{The four synthetic tasks. Inputs act from $t = 1$\,s on tasks~1 to~3 unless a time is given, and from $t = 0$ on task~4. On task~4, $J_{\mathrm{eq}}$ and $d_{\mathrm{eq}}$ refer rotor and load to the motor shaft through the ratio. The last column gives the quantities that $V_2$ compares and their bands.}
\label{tab:tasks}
\begin{tabular}{@{}l>{\raggedright\arraybackslash}p{0.19\textwidth}>{\raggedright\arraybackslash}p{0.245\textwidth}>{\raggedright\arraybackslash}p{0.25\textwidth}>{\raggedright\arraybackslash}p{0.095\textwidth}>{\raggedright\arraybackslash}p{0.1\textwidth}@{}}
\toprule
Task & Reference & Values & Experiments & Noise & $V_2$ \\
\midrule
1 & \texttt{Force} on \texttt{Mass}, \texttt{SpringDamper} to \texttt{Fixed}; position and speed sensors &
  $m = 2.0$\,kg, $c = 8.0$\,N/m, $d = 0.8$\,N\,s/m &
  fit: pulse of $5$\,N for $0.5$\,s; step of $2$\,N. Validation: sine of $3$\,N at $0.3$\,Hz. $25$\,s at $0.05$\,s, seed 42 &
  $y_s$: $3$\,mm; $y_v$: $10$\,mm/s &
  1\,\% on $m$, $c$, $d$ \\
\addlinespace
2 & \texttt{Force} on \texttt{MassWith\-Stop\-And\-Friction}, \texttt{Spring} to \texttt{Fixed}; position sensor &
  $m = 1.5$\,kg, $c = 5.0$\,N/m, $F_C = 0.6$\,N, $F_{\mathrm{prop}} = 0.3$\,N\,s/m; no Stribeck term, stops at $\pm 100$\,m &
  fit: pulse of $8$\,N for $0.5$\,s; steps of $0.5$ and $3$\,N. Validation: sine of $4$\,N at $0.25$\,Hz. $25$\,s at $0.05$\,s, seed 4242 &
  $y_s$: $2$\,mm &
  5\,\% on $m$, $c$, $F_C$, $F_{\mathrm{prop}}$ \\
\addlinespace
3 & \texttt{SignalVoltage}, \texttt{Resistor}, \texttt{Inductor}, \texttt{RotationalEMF} in series; \texttt{Inertia} with \texttt{Damper} to \texttt{Fixed}; current and speed sensors &
  $R = 2.0\,\Omega$, $L = 1.0$\,mH, $k = 0.1$\,N\,m/A, $J = 2\cdot10^{-3}$\,kg\,m$^2$, $d = 5\cdot10^{-4}$\,N\,m\,s/rad &
  fit: step of $12$\,V at $1.013$\,s; step of $6$\,V at $1.013$\,s, off at $12.017$\,s; sine of $8$\,V at $0.2$\,Hz. Validation: sine of $10$\,V at $1$\,Hz. $25$\,s at $0.05$\,s, seed 333 &
  $y_i$: $0.01$\,A; $y_w$: $0.05$\,rad/s &
  2\,\% on $k$, $J$, $d$; 5\,\% on $R$; factor 3 on $L$ \\
\addlinespace
4 & \texttt{SignalVoltage} from a \texttt{CombiTable1Ds} over the integrated state of charge; two \texttt{Resistor}s, \texttt{Inductor}, \texttt{RotationalEMF}; rotor \texttt{Inertia} and \texttt{Damper}, \texttt{IdealGear}, load \texttt{Inertia} and \texttt{Damper}, \texttt{Torque} as brake; current and output-speed sensors &
  $R_i = 0.14\,\Omega$, $R_a = 0.36\,\Omega$, $L_a = 0.042$\,H, $k = 1.15$\,V\,s/rad, $J_{\mathrm{rot}} = 0.014$\,kg\,m$^2$, $d_{\mathrm{rot}} = 0.0025$\,N\,m\,s/rad, ratio $5.2$, $J_{\mathrm{load}} = 4.8$\,kg\,m$^2$, $d_{\mathrm{load}} = 0.35$\,N\,m\,s/rad, $Q_{\max} = 2\cdot10^{5}$\,C from $0.85$; open-circuit voltage $198$, $216$, $226$, $236$, $246$, $253$, $260$\,V at state of charge $0$, $0.1$, $0.25$, $0.5$, $0.75$, $0.9$, $1$ &
  fit: brake torque of $50$\,N\,m over $20$\,s at $0.02$\,s; staircase of $100$, $300$, $500$, $200$\,N\,m, $30$\,s each, at $0.05$\,s; $400 \pm 200$\,N\,m sine of period $300$\,s over $1800$\,s at $0.5$\,s. Validation: staircase of $250$, $550$, $150$, $450$\,N\,m with steps at $150$, $400$, $650$\,s, over $900$\,s at $0.5$\,s. Seed 2606 &
  $y_i$: $0.5$\,A; $y_w$: $0.2$\,rad/s &
  5\,\% on $R_i + R_a$, $L_a$, $k$, $J_{\mathrm{eq}}$, $d_{\mathrm{eq}}$ \\
\bottomrule
\end{tabular}
\end{table}

The black boxes of Table~\ref{tab:baseline} are identified from the fitting experiments alone. The linear model is an output-error state-space model of order one to six with a common denominator, parametrised in continuous time and discretised exactly \citep{ljung1999system}. Its order, and a feedthrough where needed, are chosen by leaving one fitting experiment out at a time and taking the smallest order within five per cent of the best mean error. The neural model adds a ReLU network of 32 units to the identified linear model, $x^{+} = A_d x + B_d u + h\,S\,(W_2\,\mathrm{ReLU}(W_1 [x; u] + b_1) + b_2)$. The linear part is discretised exactly and held fixed, the internal step is $h = 12.5$\,ms and $S$ is the scale of the states, and the model starts at rest and outputs its first states. A penalty on the network term and the length of training are chosen as the linear order is, by leaving one fitting experiment out at a time, over penalties from 0 to 1000, three seeds and up to 2000 steps. On task~1 every correction predicts the left-out experiment worse than the linear model, so the selection returns the linear model itself. On task~2 it settles on 100 steps, and Table~\ref{tab:baseline} shows three seeds trained so on all fitting experiments. Trained without the penalty for 2000 steps, the networks reach $1.0$ times the noise on the fitting data of task~1 and $1.7$ on task~2, but they miss the validation experiment by 20 to 37 and by 279 to 587.

\section{The Features of the Magnet}
\label{app:features}

The behaviour layer on the magnet compares nine features, read by the same rules from the measured and the simulated traces (Table~\ref{tab:features}). Switch-on is the instant the current exceeds ten per cent of its plateau, switch-off the instant it falls below half of it after the plateau. The magnet counts as closed when the gap is below $30\,\mu$m and as open when it exceeds $0.5$\,mm. The plateau is averaged from $7$ to $10.5$\,ms in the voltage-controlled experiment and from $15$ to $22$\,ms in the current-controlled ones. The peak closing speed is the extreme velocity between switch-on and first contact. The list and its tolerances were fixed after the first run of the base cell; the reading of the speed was later made precise without changing any value on the measurement. The frozen list also records the static force at the stroke for $4$ and $8$\,A, two saturation ratios and five qualitative signatures, among them the mechanical bounce after the first impact and the dip in current during voltage-controlled closing, induced as the closing gap changes the inductance of the magnetic circuit. The static quantities are not graded because the required interface carries no mechanical flange at which the evaluator could hold the armature at a given gap. The qualitative signatures are reported beside the gate and do not enter it. On the magnet, $V_2$ compares the number of turns and the spring rate within one per cent of their stated values, and the moving mass and the coil resistance within five per cent of their measured bands, $18.6$ to $21.2$\,g and $0.78$ to $0.80\,\Omega$.

\begin{table}[tp]
\centering
\small
\caption{The nine graded features of task~5 with their values on the measurement and the tolerances fixed after the first run of the base cell and before any further run. The third experiment is the validation experiment. Times are counted from switch-on or switch-off as defined in the text.}
\label{tab:features}
\begin{tabular}{@{}llrr@{}}
\toprule
Feature & Experiment & Measured & Tolerance \\
\midrule
Closing time & voltage-controlled, $6$\,V & $3.90$\,ms & $\pm 20$\,\% \\
Closing time & current-controlled, $4$\,A & $2.64$\,ms & $\pm 20$\,\% \\
Closing time & current-controlled, $8$\,A & $1.06$\,ms & $\pm 20$\,\% \\
Release time & current-controlled, $4$\,A & $1.68$\,ms & $\pm 25$\,\% \\
Release time & current-controlled, $8$\,A & $1.76$\,ms & $\pm 25$\,\% \\
Plateau current & voltage-controlled, $6$\,V & $6.82$\,A & $\pm 5$\,\% \\
Peak closing speed & voltage-controlled, $6$\,V & $-1.01$\,m/s & $\pm 30$\,\% \\
Peak closing speed & current-controlled, $4$\,A & $-1.08$\,m/s & $\pm 30$\,\% \\
Peak closing speed & current-controlled, $8$\,A & $-1.68$\,m/s & $\pm 30$\,\% \\
\bottomrule
\end{tabular}
\end{table}

\section{Sensitivity and the Recurring Optima}
\label{app:resolution}

If one reference quantity at a time is perturbed by one per cent and the result is scored as $V_3$ does, each experiment returns a residual, in units of the noise, for that error (Table~\ref{tab:perturbation}). At the reference values every experiment lies between $0.98$ and $1.06$, so a value there means the experiment does not see the perturbation. At the tolerance of five per cent, the damping of task~4 reaches no more than $1.06$ on any experiment, so a task-4 model can carry that error and pass $V_3$. For the two mechanisms that stay open on task~4, we fitted the reference structure and the same structure with a Coulomb torque in place of both dampers to the three fitting experiments, with the same seven free quantities. Both reach at most $1.03$ times the noise on every issued experiment, and from six random starts the Coulomb fit stays about 300 above the reference structure in $\chi^2$ over $14\,006$ residuals.

\begin{table}[t]
\centering
\caption{Sensitivity of the four synthetic tasks to one parameter at a time: residual in units of the noise when one reference quantity is perturbed by one per cent, the larger of the two signs, per experiment. The other quantities stay at their reference values, and of two outputs the worse is taken. A value near one means that the experiment does not see the perturbation. Task~1 has two fitting experiments. The last row of task~3 perturbs the inductance by the factor of three that $V_2$ admits.}
\label{tab:perturbation}
\footnotesize
\setlength{\tabcolsep}{3.5pt}
\begin{tabular}[t]{@{}lcccc@{}}
\toprule
Perturbed quantity & exp1 & exp2 & exp3 & validation \\
\midrule
\multicolumn{5}{@{}l}{\textit{Task 1}} \\
mass $m$                         & 1.88 & 1.22 & -- & \textbf{12.88} \\
stiffness $c$                    & 1.91 & 1.48 & -- & \textbf{14.19} \\
damping $d$                      & 1.04 & 1.03 & -- & \textbf{2.82} \\
\addlinespace
\multicolumn{5}{@{}l}{\textit{Task 2}} \\
mass $m$                         & 2.71 & 1.01 & 1.09 & \textbf{28.35} \\
stiffness $c$                    & 3.17 & 1.01 & 3.36 & \textbf{35.07} \\
viscous $F_{prop}$               & 1.54 & 1.01 & 1.06 & \textbf{3.89} \\
Coulomb $F_C$                    & 2.97 & 1.01 & 2.16 & \textbf{2.48} \\
\bottomrule
\end{tabular}\hfill
\begin{tabular}[t]{@{}lcccc@{}}
\toprule
Perturbed quantity & exp1 & exp2 & exp3 & validation \\
\midrule
\multicolumn{5}{@{}l}{\textit{Task 3}} \\
resistance $R$                   & 2.74 & 1.55 & 4.00 & \textbf{4.79} \\
constant $k$                     & 17.18 & 6.01 & 7.76 & \textbf{5.04} \\
inertia $J$                      & 1.76 & 1.39 & 3.92 & \textbf{4.78} \\
damping $d$                      & 2.21 & 1.22 & 1.24 & \textbf{1.04} \\
inductance $L$                   & 1.06 & 1.00 & 0.99 & \textbf{1.02} \\
inductance $L$, factor 3         & 1.12 & 1.04 & 1.11 & \textbf{3.04} \\
\addlinespace
\multicolumn{5}{@{}l}{\textit{Task 4}} \\
resistance $R_i + R_a$           & 1.06 & 1.02 & 1.06 & \textbf{1.02} \\
inductance $L_a$                 & 1.07 & 1.02 & 1.01 & \textbf{1.02} \\
constant $k$                     & 2.25 & 1.97 & 1.81 & \textbf{1.78} \\
inertia $J_{eq}$                 & 1.10 & 1.03 & 1.01 & \textbf{1.02} \\
damping $d_{eq}$                 & 1.02 & 1.01 & 1.01 & \textbf{1.02} \\
capacity $Q_{max}$               & 1.02 & 1.01 & 1.02 & \textbf{1.02} \\
\bottomrule
\end{tabular}
\end{table}

The recurring optima of task~1 and task~2 in Section~\ref{sec:results} are not local minima of the objective defined by the reference experiments. Starting from a point representative of the eight task-2 runs, a trust-region least-squares fit and a Nelder--Mead search, both driven by the input as it was applied, converge to the reference values, and $\chi^2$ over the fitting experiments falls from $15\,902$ to $1\,461$. All 16 runs of the two groups interpolate the samples linearly. Under this input the best fit of the reference structure, reached from the reference values and from the runs' values alike, lies at $-2.48$, $-0.44$, $-2.02$ and $+5.28$ per cent in mass, stiffness, Coulomb force and viscous coefficient, against $-2.47$ to $-2.64$, $-0.51$ to $-0.62$, $-2.35$ to $-2.67$ and $+5.8$ to $+6.7$ per cent in the runs. Its residual is $3.7$ times the noise on the first fitting experiment and $55$ on the validation experiment. On task~1 the same reconstruction places the best fit at a mass $1.05$ per cent low, where eight runs lie between $1.04$ and $1.06$ per cent low. The reconstruction thus places the best fit close to the values the runs share, in direction and in size, so the optima are plausibly a property of the objective the agents build from the handover rather than of one LLM.

\section{Assemblies at the Noise Floor}
\label{app:assemblies}

Table~\ref{tab:assemblies} lists the component assemblies among the runs that reach the noise floor, counted as in Equation~\ref{eq:solutionset}, and the mechanisms they realise are separated in Section~\ref{sec:results}. On task~1 the two assemblies differ only in whether a \texttt{SpringDamper} or a spring and a damper in parallel carry elasticity and damping. On task~3 a single run makes up the first row. It builds the converter from a controlled voltage and a controlled torque whose two gains read one shared constant, and the damping from a \texttt{LinearSpeedDependentTorque} against the housing. Run side by side with the reference structure on the same parameters, it agrees with the reference within $10^{-7}$ of the noise on every issued experiment, also at three times the input amplitude and for a second set of parameters. The model that creates energy, which rebuilds the converter in the same way but with two separately fitted constants, fails conformance at its battery and is therefore not in the table.

The invariant $V_4$ was computed for all 68 models with a converter that the evaluator can simulate, on the first fitting experiment of their task. Of the four that rebuild the converter, three declare one constant for both directions, and the fourth is the model that creates energy. To find other places where a model could create energy, we read all 162 models of the base matrix that the evaluator can simulate through the compiler's interface and listed every physical source driven from inside the model. Besides the converters, these are six batteries on task~4 whose voltage follows a charge the model counts itself and two magnet models whose reluctances feed back their own flux, and all of them close their balance to $1.0000$. On the magnet, one model clips the length of its force-producing flux tubes at the end stop and violates the balance by $1.2\cdot10^{-4}$ of the mechanical work, far below anything a trace could show.

\begin{table}[t]
\centering
\caption{Component assemblies among the runs that reach the noise floor. Sensors, sources, ground and fixed reference points are left out because they are measurement rather than mechanism, and signal-path arithmetic (gains, sums, constants) because it is notation rather than mechanism. Two runs that differ only in how they read out or scale a signal therefore count as one assembly, and a converter rebuilt from controlled sources and gains appears without a converter component. \emph{Solved} is the number of runs at the noise floor that the assemblies are drawn from.}
\label{tab:assemblies}
\footnotesize
\begin{tabular}{@{}ll>{\raggedright\arraybackslash}p{0.82\textwidth}@{}}
\toprule
Task & Solved & Assemblies \\
\midrule
1 & 19 & \texttt{Mass}, \texttt{SpringDamper} \\
  &    & \texttt{Damper}, \texttt{Mass}, \texttt{Spring} \\
\addlinespace
2 & 19 & \texttt{MassWithStopAndFriction}, \texttt{Spring} \\
  &    & \texttt{MassWithStopAndFriction}, \texttt{SpringDamper} \\
  &    & \texttt{Damper}, \texttt{MassWithStopAndFriction}, \texttt{Spring} \\
  &    & \texttt{Mass}, \texttt{SpringDamper}, \texttt{SupportFriction} \\
  &    & \texttt{Damper}, \texttt{Mass}, \texttt{Spring}, \texttt{SupportFriction} \\
\addlinespace
3 & 20 & \texttt{Inductor}, \texttt{Inertia}, \texttt{LinearSpeedDependentTorque}, \texttt{Resistor} \\
  &    & \texttt{Damper}, \texttt{Inductor}, \texttt{Inertia}, \texttt{Resistor}, \texttt{RotationalEMF} \\
\addlinespace
4 & 22 & \texttt{Capacitor}, \texttt{DC\_PermanentMagnet}, \texttt{IdealGear}, \texttt{Resistor} \\
  &    & \texttt{CellStack}, \texttt{DC\_PermanentMagnet}, \texttt{IdealGear}, \texttt{Inertia}, \texttt{SignTorque} \\
  &    & \texttt{BearingFriction}, \texttt{Capacitor}, \texttt{IdealGear}, \texttt{Inductor}, \texttt{Inertia}, \texttt{Resistor}, \texttt{RotationalEMF} \\
  &    & \texttt{Capacitor}, \texttt{Damper}, \texttt{IdealGear}, \texttt{Inductor}, \texttt{Inertia}, \texttt{Resistor}, \texttt{RotationalEMF} \\
  &    & \texttt{Capacitor}, \texttt{IdealGear}, \texttt{Inductor}, \texttt{Inertia}, \texttt{Resistor}, \texttt{RotationalEMF}, \texttt{SignTorque} \\
  &    & \texttt{CellStack}, \texttt{Damper}, \texttt{IdealGear}, \texttt{Inductor}, \texttt{Inertia}, \texttt{Resistor}, \texttt{RotationalEMF} \\
  &    & \texttt{CellStack}, \texttt{IdealGear}, \texttt{Inductor}, \texttt{Inertia}, \texttt{Resistor}, \texttt{RotationalEMF}, \texttt{SignTorque} \\
  &    & \texttt{Capacitor}, \texttt{CombiTable1Ds}, \texttt{Damper}, \texttt{IdealGear}, \texttt{Inductor}, \texttt{Inertia}, \texttt{Resistor}, \texttt{RotationalEMF} \\
\bottomrule
\end{tabular}
\end{table}

\section{Configurations and Evaluation Settings}
\label{app:serving}

The follow-up on the acceptance criterion ran Sonnet~5 under Claude Code 2.1.300 to 2.1.301 and Qwen3.8 under OpenCode 1.18.25.

{\emergencystretch=3em Besides the nine LLMs of Table~\ref{tab:configs}, seven ran under OpenCode: Gemma~4 (\nolinkurl{google/gemma-4-31b-it}), gpt-oss-120b (\nolinkurl{openai/gpt-oss-120b}), Mistral Medium 3.5 (\nolinkurl{mistralai/mistral-medium-3-5}), Muse Glimmer (\nolinkurl{meta/muse-glimmer-30b}), Nemotron 3 Super (\nolinkurl{nvidia/nemotron-3-super-120b-a12b}), GLM-4.7-Flash (\nolinkurl{z-ai/glm-4.7-flash}) and MiniMax M3 (\nolinkurl{minimax/minimax-m3}). All runs used OpenModelica 1.27.0 with the Modelica Standard Library 4.1.0 in the same sandbox. The evaluator connects each delivered model to the same source blocks that generated the synthetic data, so the input is exact rather than interpolated from the files, and to the recorded voltage as a linearly interpolated table on the magnet. It simulates with the default DASSL solver of OpenModelica at a relative tolerance of $10^{-8}$, with one output interval per sampling interval, two on the magnet and the default event handling, and interpolates the outputs linearly onto the measurement times before the RMSE is taken.\par}

\begin{table}[t]
\centering
\small
\setlength{\tabcolsep}{6pt}
\caption{Configurations of the nine LLMs that reach the noise floor at least once, as recorded in the run manifests: harness and its versions and number of runs, without the follow-up on the acceptance criterion. Every configuration ran at the highest reasoning setting its provider offers.}
\label{tab:configs}
\begin{tabular}{@{}llr@{}}
\toprule
Configuration & Harness & Runs \\
\midrule
Opus~5 & Claude Code 2.1.251--2.1.281 & 22 \\
Sonnet~5 & Claude Code 2.1.251--2.1.266 & 17 \\
DeepSeek V4 Pro 0813 & OpenCode 1.18.19--1.18.25 & 17 \\
gpt-5.6-luna & Codex 0.147.0 & 21 \\
gpt-5.6-sol & Codex 0.147.0--0.154.0 & 22 \\
gpt-5.6-terra & Codex 0.147.0 & 19 \\
Hy4 preview & OpenCode 1.18.25 & 19 \\
Kimi~K3 & OpenCode 1.18.19--1.18.25 & 28 \\
Qwen3.8 (27.8\,B) & OpenCode 1.18.19--1.18.25 & 37 \\
Qwen3.8 (27.8\,B) & Claude Code 2.1.284--2.1.285 & 3 \\
Qwen3.8 (27.8\,B) & Codex 0.154.0 & 3 \\
\bottomrule
\end{tabular}
\end{table}

\section{Reading of Seven Delivered Models}
\label{app:expert-models}

The builder read seven delivered models from six of the nine configurations, two of them from Opus~5, after rating the charts. The reading covered the electrical, magnetic and mechanical circuits and the diagram as delivered. Parameter values are quoted from the delivered \texttt{model.mo} files. Measured reference values are the pole radii $4.9$ and $11.5$\,mm, the armature and spring masses $18.6$ and $2.6$\,g, and the coil resistance of $0.78$ to $0.80$\,$\Omega$.

\begin{experttable}
\centering
\footnotesize
\caption{Task~5, seven delivered models read by the builder. Tier is the builder's overall rating of the run's standard chart: \protect\extier{exgreen}~good, \protect\extier{exyellow}~reasonable, \protect\extier{exorange}~not good, \protect\extier{exred}~not viable. \emph{Reuse} is whether the builder would adapt the model to a geometry variant rather than refit it; \emph{Usable} is whether the diagram as delivered could be used without redrawing it; rearranging its elements is allowed, which is why five are usable while only one could be worked with without any rearrangement.}
\label{tab:expert-models}
\setlength{\tabcolsep}{4pt}
\begin{tabular}{@{}lc>{\raggedright\arraybackslash}p{0.6\textwidth}cc@{}}
\toprule
Run & Tier & Magnetic network, assessment condensed & Reuse & Usable \\
\midrule
Opus~5, 5 & \extier{exgreen} &
  Yoke, poles and armature as geometric flux tubes; pole radii $5.35$ and
  $11.67$\,mm; eddy path in the inner pole; two leakage paths around the
  poles and one between them. Moving mass $21.18$\,g, coil $0.786$\,$\Omega$.
  Corner treatment inconsistent between inner and outer pole; outer fringing
  radius $8.6$\,mm larger than the geometry supports. ``Would pass with good
  marks.'' & yes & yes \\
\addlinespace
Opus~5, 1 & \extier{exgreen} &
  The network of run~5 element for element, in almost the same arrangement;
  pole radii $5.77$ and $11.72$\,mm; yoke bottom fitted at $1.05$\,mm where
  run~5 fixed it at $3$; one fringing radius of $1.4$\,mm for both poles.
  Moving mass $19.60$\,g, coil $0.793$\,$\Omega$. A damper of $0.20$\,Ns/m
  on the armature; stiffness and damping set separately for the two end
  stops, $278$ and $83$\,kN/mm. ``Comparable to run~5, if anything more
  clearly organised.'' & yes & yes \\
\addlinespace
Kimi~K3, 1 & \extier{exgreen} &
  Two lumped iron elements with fitted length and area; one working gap of
  radius $4.17$\,mm, the outer pole absent; fringing width $68$\,mm without
  geometric meaning. Remanence as a magnetomotive force of $35.6$\,A.
  Series reluctance equivalent to $102\,\mu$m of air beside a mechanical
  residual gap of $5\,\mu$m: the same effect twice. ``Clearly
  inferior'' as a network. & no & yes \\
\addlinespace
Hy4, 1 & \extier{exgreen} &
  Two working gaps with geometric radii $5.44$ and $11.25$\,mm; one leakage
  permeance. Iron as a self-written tabulated flux tube with
  $\mu_r = 1 + |B|/(\mu_0 H)$, the constant being an error; no iron in the
  armature; no eddy path. Diagram ``dramatically bad''. & partly & no \\
\addlinespace
sol, 1 & \extier{exyellow} &
  Coil, magnetic circuit and armature as separate classes. One working gap
  from a fitted area of $43.8$\,mm$^2$ without physical meaning; fringing
  dimensions fitted, not derived. Parasitic inductance of $47\,\mu$H on
  the electrical side. Stribeck friction, alone among the seven. Its stop
  parameters are round numbers and its two stops share a stiffness, which
  the reviewer reads as a structure left uncalibrated. & no & yes \\
\addlinespace
DeepSeek, 3 & \extier{exyellow} &
  One working gap, one iron element of length $38.8$\,mm and area
  $69.9$\,mm$^2$, close to the device; no fringing, no leakage, no eddy path.
  A mechanical damper of $49.7$\,Ns/m stands in for the eddy losses, as the
  agent's comments state. ``Simple and easy to understand.'' & no & yes \\
\addlinespace
luna, 3 & \extier{exred} &
  Two hand-assembled branches of library components, slow and fast, in parallel; iron as a variable
  reluctance driven from table data; leakage inductance of $316\,\mu$H
  on the electrical side. Not a network in the sense of the task. & no & no \\
\bottomrule
\end{tabular}
\end{experttable}

\section{Framing Instruction}
\label{app:prompt-frame}

Every run received two texts and nothing else: the framing instruction below, identical for every run, and the task description in its working directory. The framing instruction is given as the agents received it.

{\microtypesetup{activate=false}%
\setlength{\topsep}{0pt}\setlength{\partopsep}{0pt}%
\begin{Verbatim}[fontsize=\fontsize{9}{9.3}\selectfont,fontfamily=lmtt,breaklines=true,breakanywhere=true]
You are asked to build and calibrate a Modelica model. The task is defined in
`task.md` in your working directory. Read it first and follow it exactly.

**Working directory.** Do all your work inside `work/`. Run every OpenModelica
build inside `work/build/`, because `omc` writes its build artefacts into the
current directory. Your deliverables (`model.mo`, the report, any scripts you
write) belong directly in `work/`.

**Tools.** `omc` is available system-wide, the Modelica Standard Library 4.1.0
is installed. A Python environment with numpy, scipy and OMPython is at
`.venv/bin/python` relative to the project root. You have network access; use
it for documentation if you need it.

**Run everything synchronously.** Never end your turn while a job is still
running. Do not start background jobs, do not detach processes, do not poll for
results in a later turn. If a calibration or simulation takes long, run it in
the foreground and wait for it to finish within the same turn, even if that
takes several minutes. Ending a turn while waiting is the single most common
way this workflow fails.

**Do not guess class names.** The Modelica Standard Library changed names
between major versions. Verify every class you use by introspection, for
example with `getClassNames`, instead of relying on memory.

**Check that parameters are actually bound.** OpenModelica silently fills
missing parameter values with start values and only emits a warning, so a model
can compile and simulate while being physically dead. Verify that every
parameter you intend to fit is bound to a value you set.

**Information boundaries.** Everything you may use is inside your working
directory. There is no other material for you to find, and looking for
reference solutions elsewhere on the system or on the network is not part of
the task. The environment enforces this, so this note is a statement of intent,
not a restriction you need to work around.

**Report.** Deliver the output package listed in `task.md`. Be explicit about
assumptions you added that the description does not state, and about
alternatives you considered and rejected.
\end{Verbatim}
}

\end{document}